\documentclass[12pt,letterpaper]{article}
\usepackage{times}
\usepackage{epsfig}
\usepackage{rotating}
\usepackage{subfig}
\usepackage{graphicx}
\usepackage{tabularx}
\usepackage{color}
\usepackage{array}
\usepackage[margin=0.75in]{geometry}
\usepackage{amsmath,amsthm, amssymb, latexsym, enumerate}

\newtheorem{theorem}{Theorem}

\theoremstyle{remark}
\newtheorem*{remark}{Remark}

\theoremstyle{definition}
\newtheorem{definition}{Definition}[section]
\newtheorem{corollary}{Corollary}[section]

\usepackage{mathtools}

\begin{document}

\title{On Qualitative Shape Inferences: \\
a journey from geometry to topology}
\author{Steven W. Zucker\thanks{The paper describes research done in collaboration with O. ben-Shahar, P. Breton, D. Holtmann-Rice, and B. Kunsberg. We thank E. Connor, A. Gyulassy, and J. Todd for permission to use figures. It was supported by the Paul G. Allen Frontiers Group, the Simons Collaboration on the Global Brain, and NSF CRCNS AWD0001607.}\\
Computer Science \& Biomedical Engineering \\
Yale University \\
New Haven, CT}
\maketitle

\abstract{Shape inference is classically ill-posed, because it involves a map from the (2D) image domain to the (3D) world. Standard approaches regularize this problem by either assuming a prior on lighting and rendering or restricting the domain, and develop differential equations or optimization solutions. While elegant, the solutions that emerge in these situations are remarkably fragile. We exploit the observation that people infer shape qualitatively; that there are quantitative differences between individuals. The consequence is a topological approach based on critical contours and the Morse-Smale complex. This paper provides a developmental review of that theory, emphasizing the motivation at different stages of the research.
}
 
\section{Introduction}

\begin{figure}[ht]
\begin{center}
\includegraphics[width=.24 \linewidth]{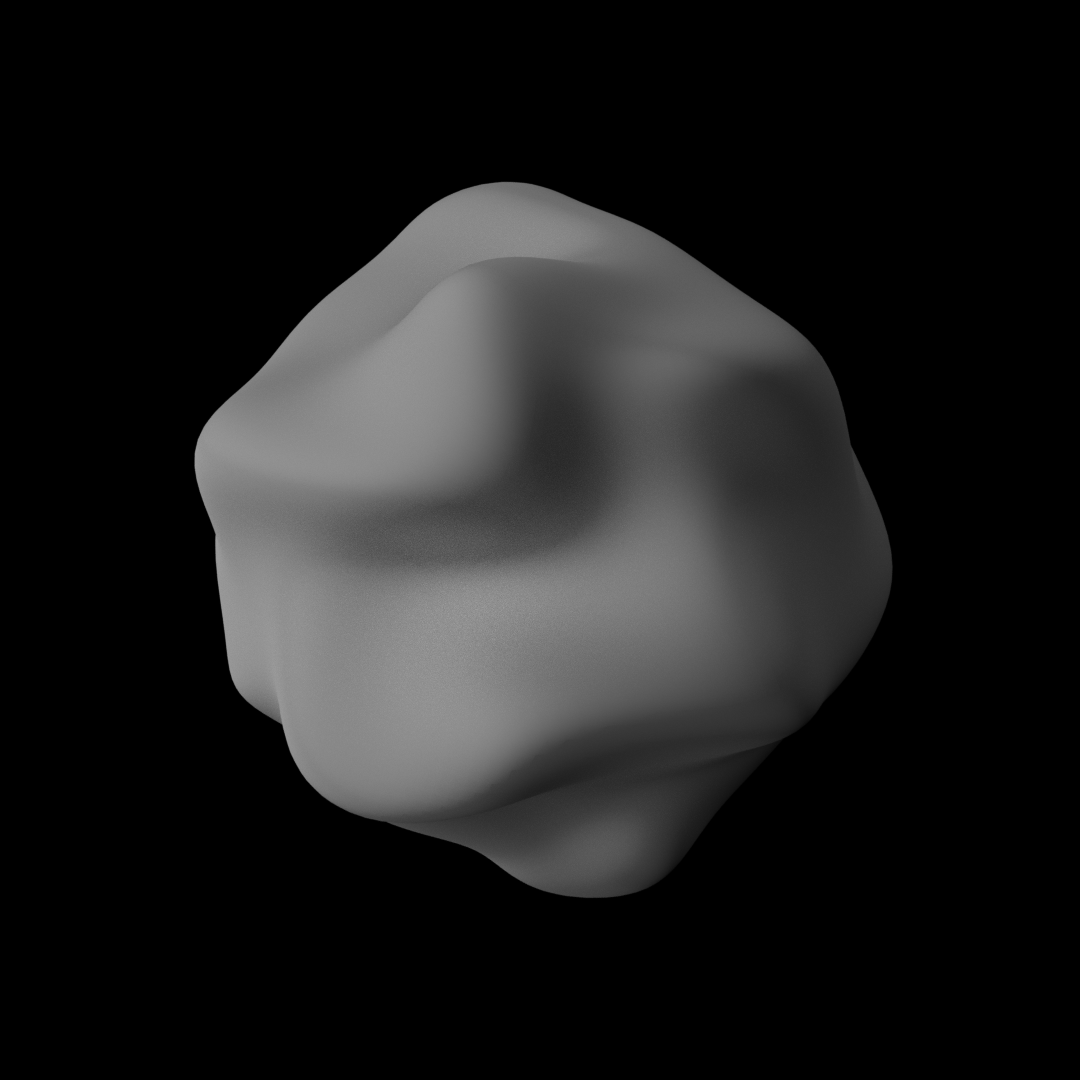}
\includegraphics[width=.24 \linewidth]{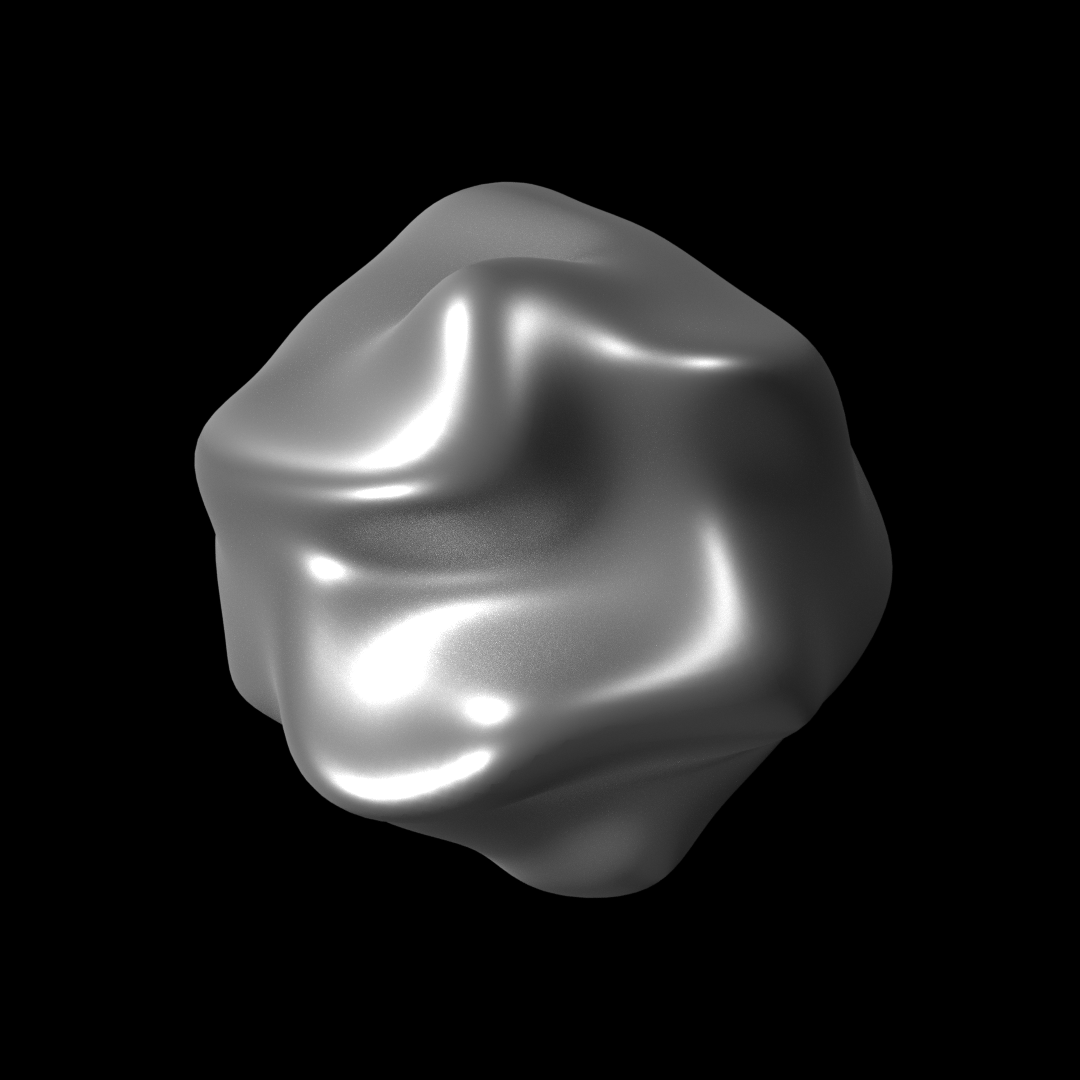}
\includegraphics[width=.24 \linewidth]{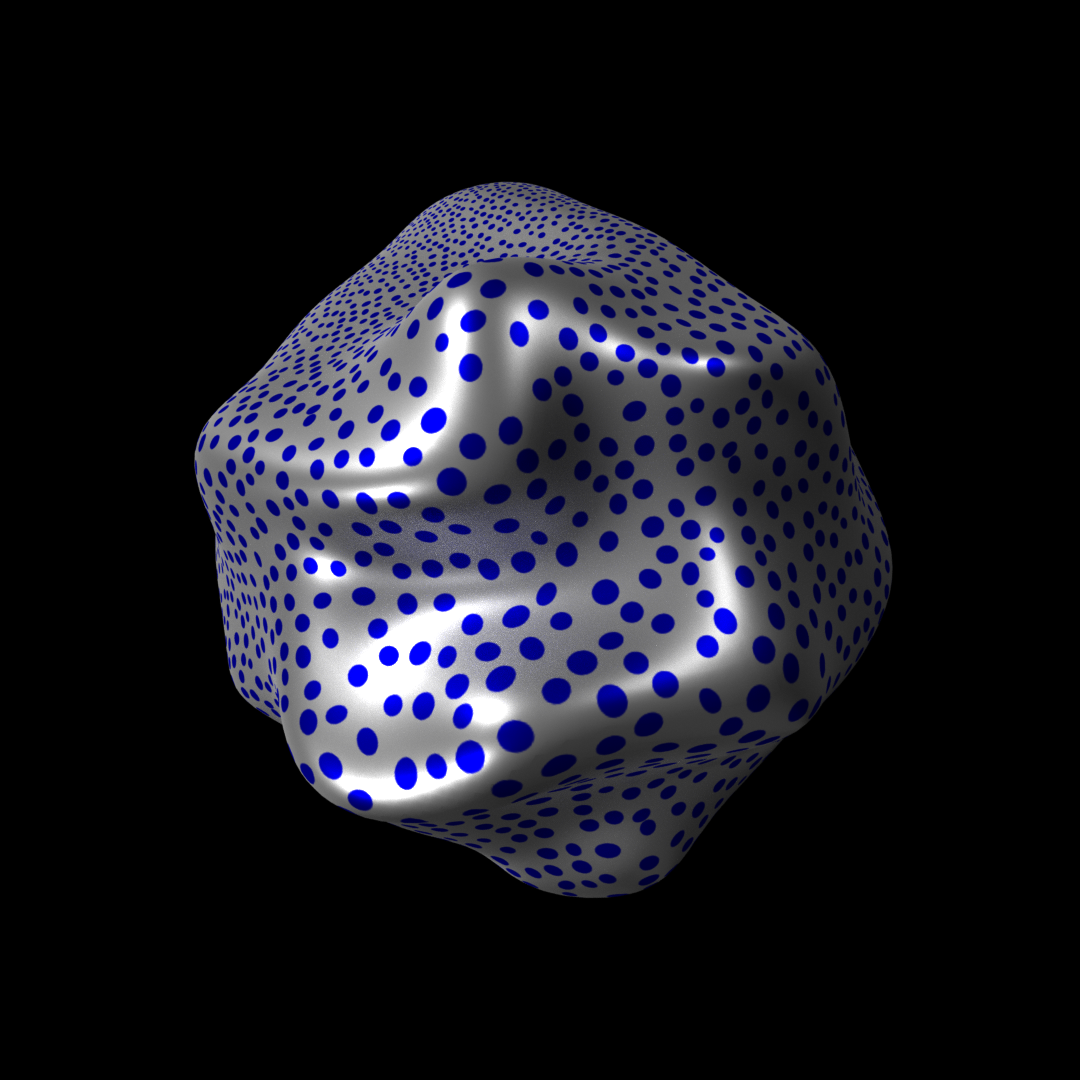}
\includegraphics[width=.24 \linewidth]{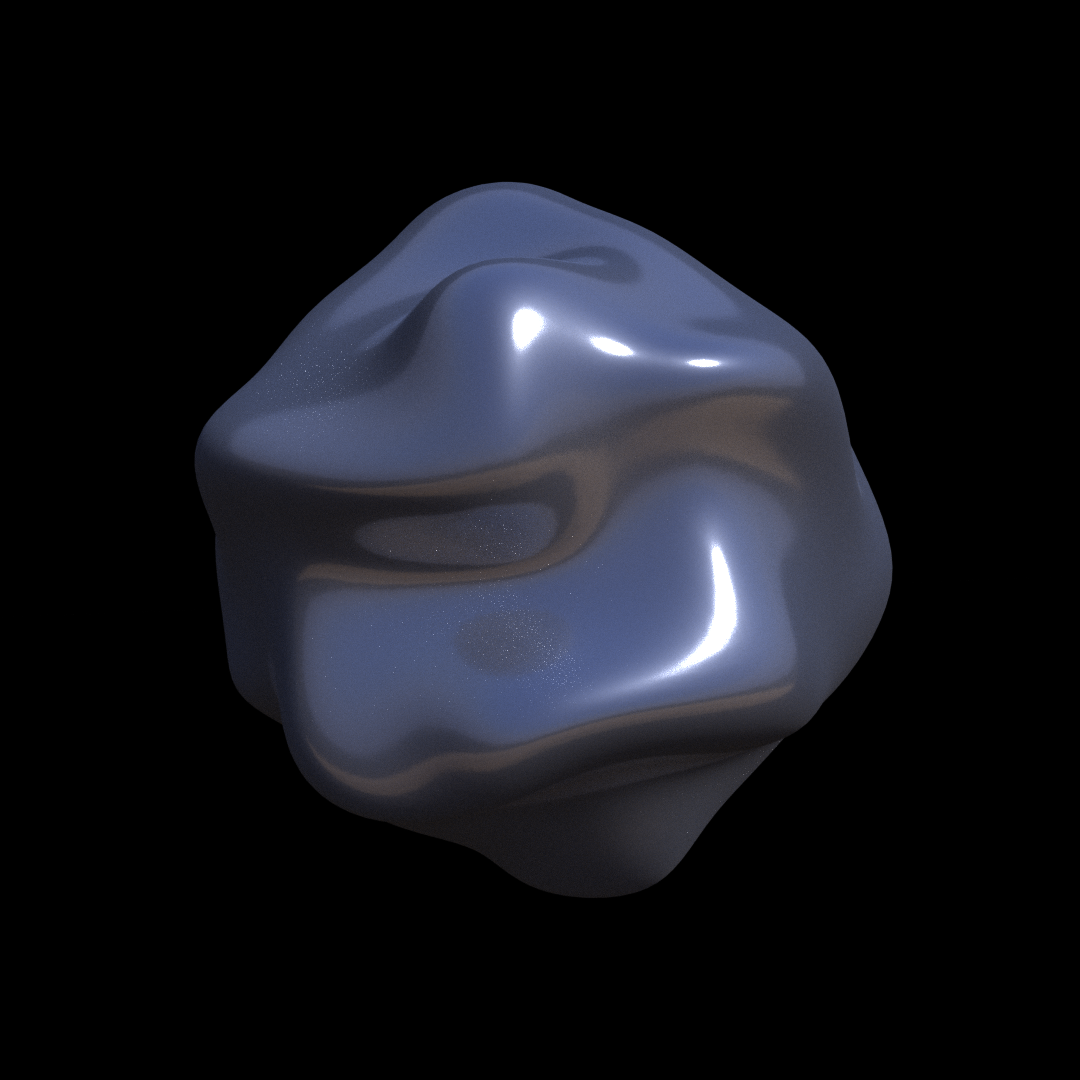}
\caption{The same shape can be rendered under a multitude of rendering functions, yielding a infinitely diverse set of images.  From left to right, we have: Lambertian shading, specular shading, texture and specular shading, and a ``Savannah" environment map.  With this in mind, how does the visual system achieve 3D shape constancy?  Here, we motivate the case for invariant features ({\em critical contours}) that delineate the protrusion boundaries. \label{fig:variety}}
\end{center}
\end{figure}

How is it possible that we are able to infer 3D shape under the infinite variety of renderings, materials and illuminations that the real world offers us? Ernst Mach \cite{mach} formalized this in what is (almost certainly) the first version of the image irradiance equation \cite{horn89}. He identified the fundamental problem: Even with orthographic projection and the light source above, the solution is "...indeterminate; i.e., many curved surfaces may correspond to one light surface [image] even if they are illuminated in the same manner" (p. 286). This ill-posedness of the image irradiance p.d.e. raises a conundrum: somehow, we effortlessly "solve" this extremely difficult mathematical problem, almost regardless of the material or lighting or the rendering function. As this example illustrates (Fig.~\ref{fig:variety}), the image differs substantially almost everywhere, yet our shape percept does not. Our long term goal is to understand how brains can achieve this.

More than a century and a half later, we are still confronting this conundrum. Mach specialized to simpler scenes (e.g., cylinders). Now researchers specialize to classes of smooth surfaces (e.g. locally  spherical \cite{pentland84}) and/or extend these priors to the scene (e.g., office scenes), to the lighting (e.g., from above), etc. (see, among others \cite{kersten2004object, prados2005shape, Barron:2012tt}). Deep neural networks \cite{2012arXiv1206.6445T} restrict training data, by specializing on 'faces' or 'chairs' or 'dormitory rooms' \cite{kulkarni2015deep}. For a recent review, see \cite{breuss2016perspectives}. While algorithms can function in these specialized situations, these approaches have turned out to be remarkably brittle. In addition, another problem is created in determining which priors to use or which scene class is being imaged. Although some progress has been made in generalizing from, say, {\em chairs} to {\em cars} \cite{sitzmann2019scene}, one's sense is that particular solutions can be engineered only for particular situations using these techniques. The robustness of our visual systems remains elusive.

Conceptually, Mach established the framework for thinking about the shape inference problem as a map from images (in 2D) to either surface normals or depths (in 3D), governed by the physics of reflectance. Under a Lambertian model, for example, the task becomes to infer which combination of lighting and surface normal at a point gave rise to the image intensity observed at that (projected) point. Consistency between locations is established by the p.d.e. or relaxed into an optimization or regularization. 

This 'inverse physics' tradition has consequences.
First, recognizing that it is difficult to generalize Lambertain to other reflectance models, it is tempting to postulate separate modules instead, as in the classical shape-from-X work in computer vision and visual psychophysics \cite{buelthoff1991shape} (X, for example, could be a physical reflectance model \cite{schlick1994survey}, a texture \cite{gaarding1992shape}, specularities \cite{adato2007toward} or focus \cite{nayar1994shape}). Each module is governed by a (differential) equation relating images to surfaces, or an optimization where the reflectance model is explicit. Second, the 'inverse physics' tradition seeks a solution that is dense, e.g. the depth or surface normal at every projected point. This also seems consistent with our own perceptual experience, in which we are able to describe the surface normal at every point \cite{Gibson52}. And third, it seeks to recover the unique solution governed by the (differential) equation. From a mathematical perspective, each of these modeling presumptions seems reasonable, although it does create the problem of integrating the separate sources of information together, a challenge in itself \cite{johnston1993integration, landy1995measurement, clark2013data}.

Current work remains largely consistent with this tradition. In the neural network community, modularization is introduced via training data although the "X's" are much finer grain \cite{kulkarni2015deep}. In some cases, the "physics engine" is actually explicit \cite{wu2015galileo}. Nevertheless, dense normal reconstructions are still sought \cite{wang2020vplnet}, and researchers in psychophysics have developed clever probes, for depth \cite{todd1989ordinal} or surface normal \cite{Koenderink92Surface}, to reveal it. The results of from these "perceptual fits" will be important later in this paper.

Line drawings raise problems for dense reconstructions, because the given information is so sparse \cite{koenderink1984does, Cole:2009:HWD}. They emphasize how information must be "filled-in" from the data. One option is to view drawings as another possible value for X \cite{Waltz75understandingline, barrow1981interpreting, Malik1987}. Another is do the filling-in at the image stage, before any 3D inference, and adversarial networks are being developed to do precisely this (e.g., \cite{chen2018sketchygan}). But this approach to the integration problem doesn't help: once the sketch is expanded into a full image, it places us back where we started -- the 3D shape still needs to be inferred.

We draw a different message from the perception of drawings; namely, that their "sparseness" carries over to continuous images in an informative sense. That is, we submit that there exist certain neighborhoods in images that are more important than others for 3D inferences, and drawings are an abstract representation -- a kind of concentration -- of them. If the structure defining these neighborhoods could be identified, it would then provide insight into the shape inference problem. This is plausible, since line drawings are not arbitrary. Of all the possible lines that could be drawn, artists emphasize certain overall features (ridges, bumps, valleys) \cite{nicolaides1941natural}. This is true when studied psychophysically as well \cite{phillips2003perceptual, Cole:2009:HWD}. Computationally, if the surface were known, techniques for generating drawings exist \cite{DeCarlo:2003:SCF, Judd07, Lee:2007:LDV:1276377.1276400}. The challenge is to be able to go in the other direction: to find image structures that stand in a one-to-one relationship with surface structures. Although a few results exist, most notably those of Koenderink and colleagues \cite{Koenderink82Perception, koenderink:1980bm}, these are all pointwise. They exploit mainly exploit notions from differential calculus. Extending invariances beyond a point has caused some confusion in the image processing literature, especially when attempting to define terms such as ridge \cite{eberly1994ridges, koenderink1994two}. 

Although, we maintain, certain image neighborhoods are more important (for shape inferences) than others, their arrangement also matters. Our percepts are global. Mathematically we must confront the transition from local (e.g. calculus) to global (e.g., topology), and such transitions have been largely missing from the shape inference literature. Possible components of shapes, including ridges, bumps, etc., are defined over a neighborhood (not simply at a point), and thus differ fundamentally from image singularities (e.g. points of maximal intensity). One class of global descriptors includes the skeleton and shock graph \cite{kimia1995shapes}, although this is more appropriate to matching and pose. We now move in a rather different direction.

Consider what "uniqueness" should mean.  This is typically drawn from p.d.e. theory, as described above. However, from a human perception perspective, perhaps different viewers interpret the same drawing slightly differently. This is plausible, since the drawing can be viewed as a set of constraints, or anchors, from which slightly different surfaces could be constructed (or different regularizers employed). Psychophysical measurements \cite{mamassian1998observer} have indeed confirmed this is the case. But it is even more surprising (at least to me) that there are real perceptual differences across subjects even when viewing full images (references and an example in Sec.~\ref{sec:qual}). We take these differences to be fundamental, and seek {\em qualitative but not quantitative similarity} across different viewers (or the same viewer at different times). Thus, following others, we postulate that the solution to the shape inference problem should not be the unique surface but rather a family of possible surfaces, thereby embracing ambiguity (e.g., \cite{Bel99}). We take a different tack. Following the global requirement above, we submit this suggests a topological rather than geometric solution, and this is what we shall characterize. It enables us to obtain consistent results over a range of cues and rendering models. Instead of the factorization into different X modules, we emphasize their common structure.

In this review we focus on two threads in our research, both motivated by biological and psychological considerations. The first thread is more traditional. We seek a system of differential equations that specify the geometry of possible surfaces, given an image under Lambertian rendering, for an arbitrary light source. Although this thread was in-line with p.d.e. approaches, it did emphasize more differential geometry than is normally applied. The motivation built upon the representation of image data in visual cortex, and transport operations along and across isophotes were studied. we sought commonalities among the different possible solutions. Although ill-posedness could not be escaped under the light-source relaxation, effective similarities to the solution in certain neighborhoods (e.g., around ridges) did emerge. This inspired our second thread of research, which is topological. It captures the qualitative structure in these image neighborhoods globally, intuitively like a drawing captures salient parts of an image and organizes them into a coherent whole. Taking a limit, the shading concentrates into what we define as {\em critical contours}. They are part of the Morse-Smale complex, a global topological invariant. Our main theorem shows how this extends to an invariant between images and surfaces for generic variations in lighting and rendering functions. 

There is another major difference from the classical approaches.
The Morse-Smale complex of critical contours reveals not the exact surface but a scaffold on which it can be built. Somehow the surface must be interpolated from this scaffold. In other words, the Morse-Smale complex acts as a constraint from which a full surface can be interpolated, but the interpolation process is otherwise less well specified. Different interpolations yield different surfaces, except around the Morse-Smale scaffold. It follows that the final surface is therefore not unique.  Surfaces will be quantitatively different -- but qualitatively similiar -- which is the price paid for ill-posedness. Satisfyingly, this is precisely what is observed psychophysically, when human shape percepts are analyzed. 

The paper is organized as follows. Since we are mainly interested in the biological perception of shape, we begin with two observations to set the stage. The first describes the representation of image information in visual cortex as shading (orientation) flows, and the second illustrates light-source invariance. In Sec.~\ref{sec:first-stage} we develop the map from a patch of shading flow (from the image) to a patch of surface (in $R^3$). The differential-geometric analysis captures the information available by taking an infinitesimal step along (or normal to) the shading flow. The second-order shading equations express this as image derivatives, on one side, and the shape operator, on the other. However since only the norm of the shape operator appears, ambiguity and ill-posedness are explicit and cannot be avoided. 

The insight from this analysis is developed in the transition, Sec.~\ref{sec:transition}, where the shading equations are expanded along ridges. It is here that the qualitative similarity emerges (see especially Fig.~\ref{fig:ridge_cross_section}) and leads to the second thread (Sec.~\ref{sec:thread2}). Here we illustrate the qualitative nature of shape perception, introduce critical contours and the Morse-Smale complex, and proceed to the main theorem. Several examples illustrate these complexes on smooth blobs and bumps. 

\section{Background on Neurobiology}

We begin with two brief remarks about the motivation from neurobiology, to set the stage for our modeling.

\begin{figure}[h]	
\begin{center}
\includegraphics[width=.7 \textwidth]{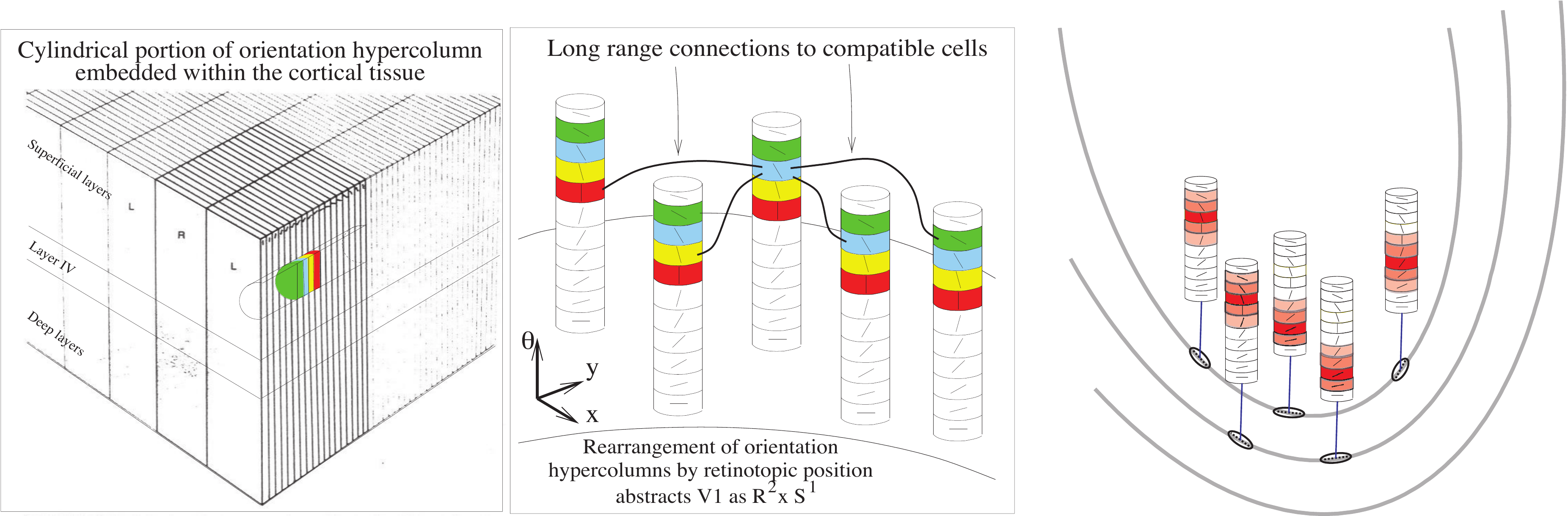} \\
\begin{tabular}{c c c}
     (a) & (b) & (c) 
\end{tabular}
\caption{Orientation selectivity and input representation. (a) A cartoon model of the first visual area. (b) Orientation hypercolumns and the long-range horizontal connections between them. (c) Orientation-selective neurons select the tangent to isophotes. }
\label{fig:biology-bundle}
\end{center}
\end{figure}

First, as is well known, visual cortex is largely organized around orientation \cite{HubelBook}; Fig.~\ref{fig:biology-bundle}. In a mathematical abstraction, V1 in primates consists of 'hypercolumns' of cells tuned to each possible orientation at each (retinotopic) location. The response of each cell is frequently modeled as a convolution against a Gabor function, followed by a non-linearity (detector).  In effect, then, the cells in V1 sample the tangent to the isophotes in a local neighborhood. We refer to this vector field as the {\em shading flow field} \cite{zucker-sff}, denoted $V(x,y)$ and observe that it defines a section in the tangent bundle to smooth images. Details in how this relates to neurobiology are described in \cite{ben2004geometrical}.

Since images are not represented directly in visual cortex, the shading flow field will serve as the basis for inferring shape. Mathematically, the shading flow field is defined via the image gradient: $V(x,y) \perp \nabla I(x,y).$ Some agree with the role of isophotes as a foundation for shape inferences, especially  \cite{koenderink:1980bm}, while others disagree \cite{doi:10.1068/i0645}. For now we shall assume it, because it sets the stage for the geometric approach, but we shall return to this issue in Sec.~\ref{sec:transition}, because it plays a large role in the transition.

\begin{figure}[ht]
\begin{center}
\begin{tabular}{c c c}
\includegraphics[width = 0.3 \linewidth]{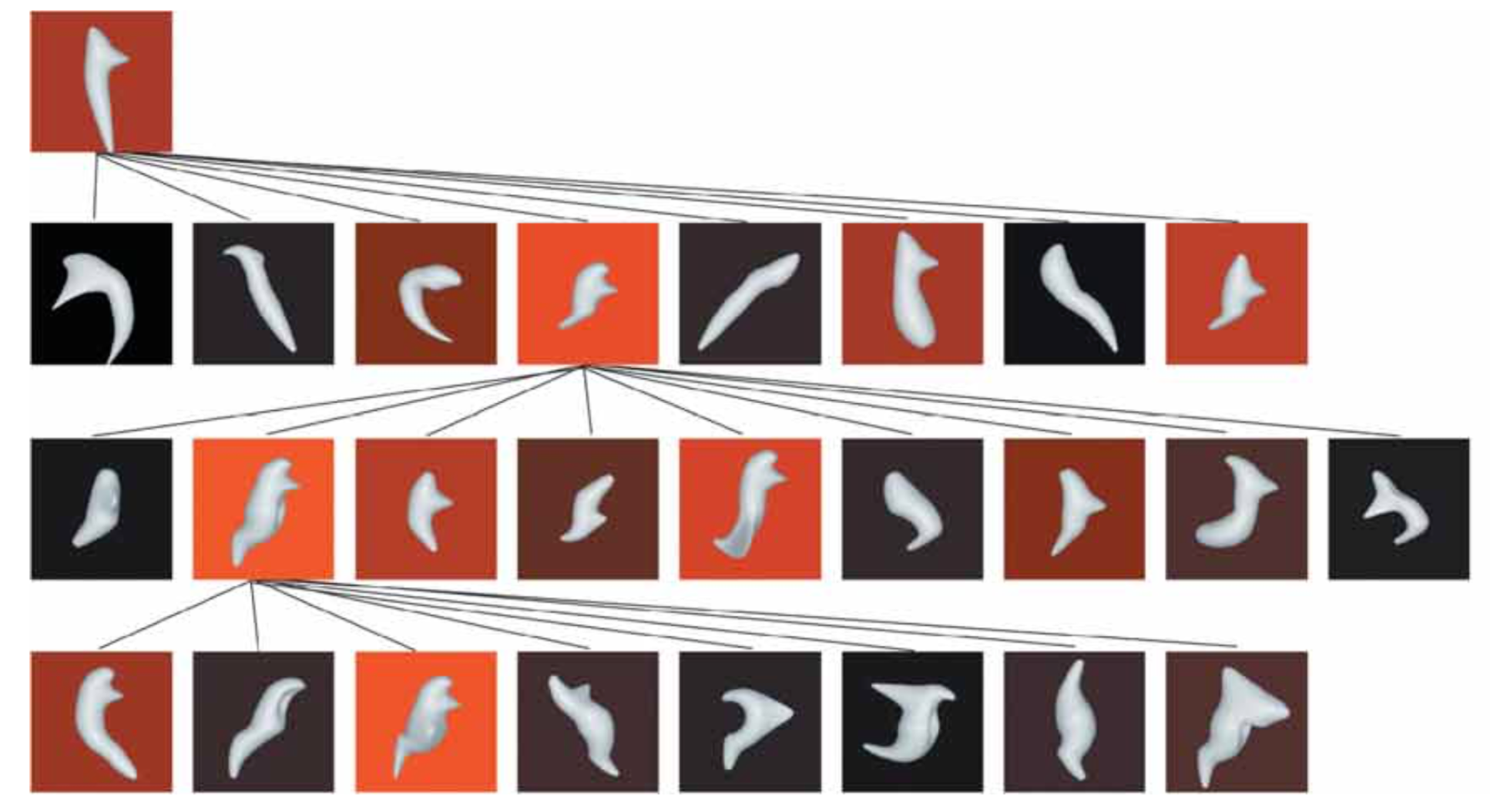} &
\includegraphics[width = 0.3 \linewidth]{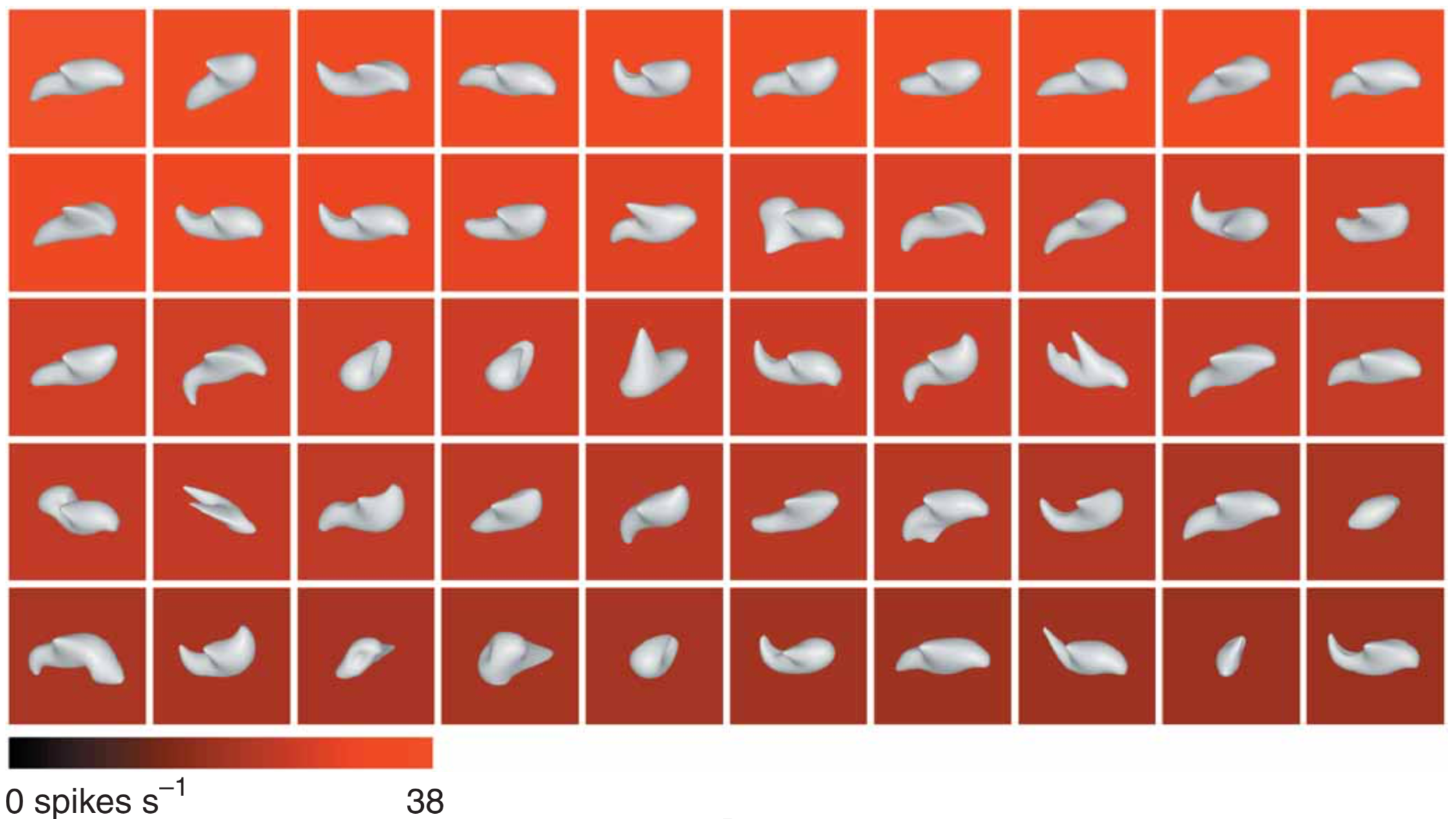} & \includegraphics[width = 0.2 \linewidth]{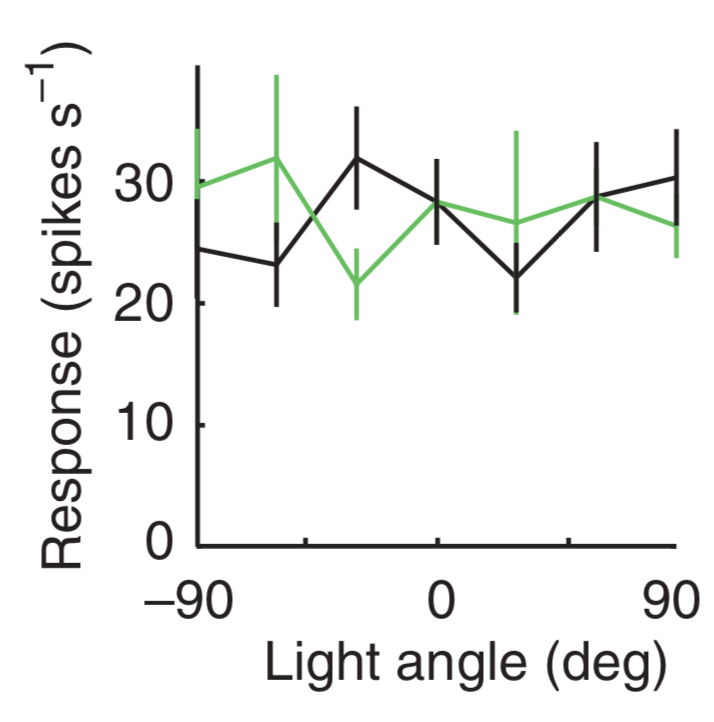} \\
(a) & (b) & (c) \\
\end{tabular}
\caption{Neurons in IT cortex respond to an assortment of different shapes but are often invariant to lighting. (a) A distribution of 2D images of 3D stimuli. The background corresponds to neural activity, with red meaning higher. (b) The preferred stimuli for one IT neuron, arranged according to strength of response. One can interpret these as containing a patch of surface to which the neuron is responsive. (c) Response invariance across lighting for two stimuli. Figures from \cite{Connor08}.} \label{fig:yamane}
\end{center}
\end{figure}

The second observation centers on higher levels of cortex, in particular the inferotemporal (IT) cortex, where (some) neurons exhibit a surprising level of invariance across shape variation and robustness across lighting. In effect, these cells are either involved in shape inference or are representing its result. Unlike IT neurons selectively tuned for e.g. faces \cite{tsao2018face}, they therefore could reveal which 'features' of a shape might matter.  Studying such cells, Yamane et al. \cite{Connor08} sought to isolate the ideal shape stimulus for an individual neuron using a genetic algorithm to search among possibile stimuli.  Examples are shown in Fig.~\ref{fig:yamane}. Two conclusions from this study are relevent. First, notice how the stimulus for one of these neurons can vary quite a bit, even though its activity remains quite high. This suggests that it is not the full shape that is driving the cell, but some part or abstraction of it. Second, changing the direction of the light source (and thus the image of the 3D object) did not change the response of the neuron. Although there are exceptions \cite{CHRISTOU19971441}), intuitively we rarely have the sense that shapes change as a function of their lighting.

Taken together, these two observations suggest the problem addressed in our first research thread.

\section{Thread 1: From Shading Flow to Surface Patch}
\label{sec:first-stage}

\begin{figure}[h]	
\begin{center}
\begin{tabular}{c}
\includegraphics[width = .8\textwidth]{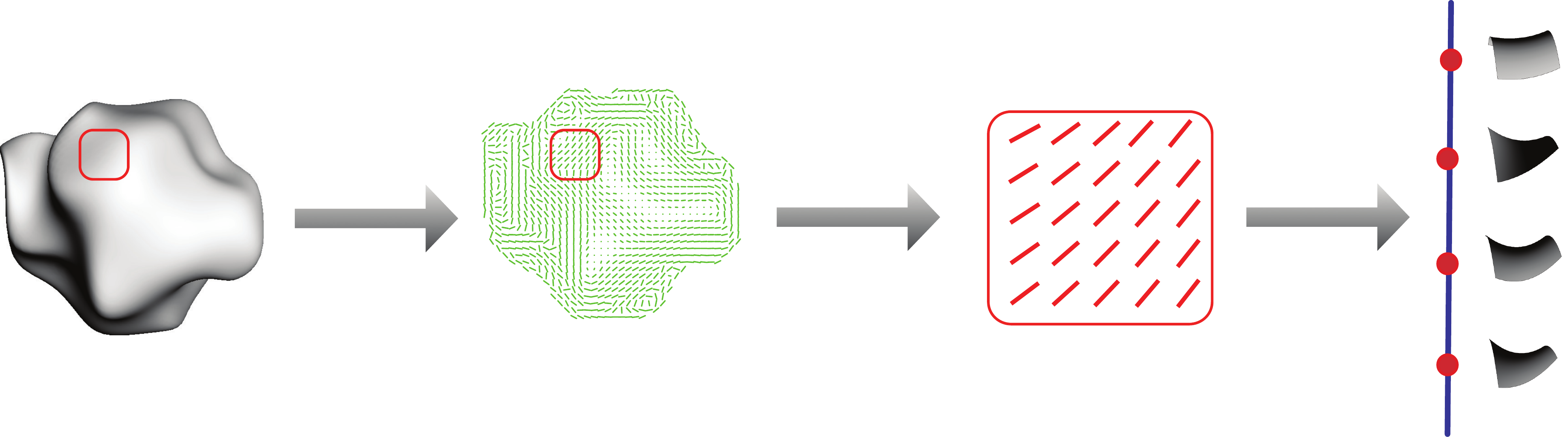} \\
(a) \\
\includegraphics[width = .8\textwidth]{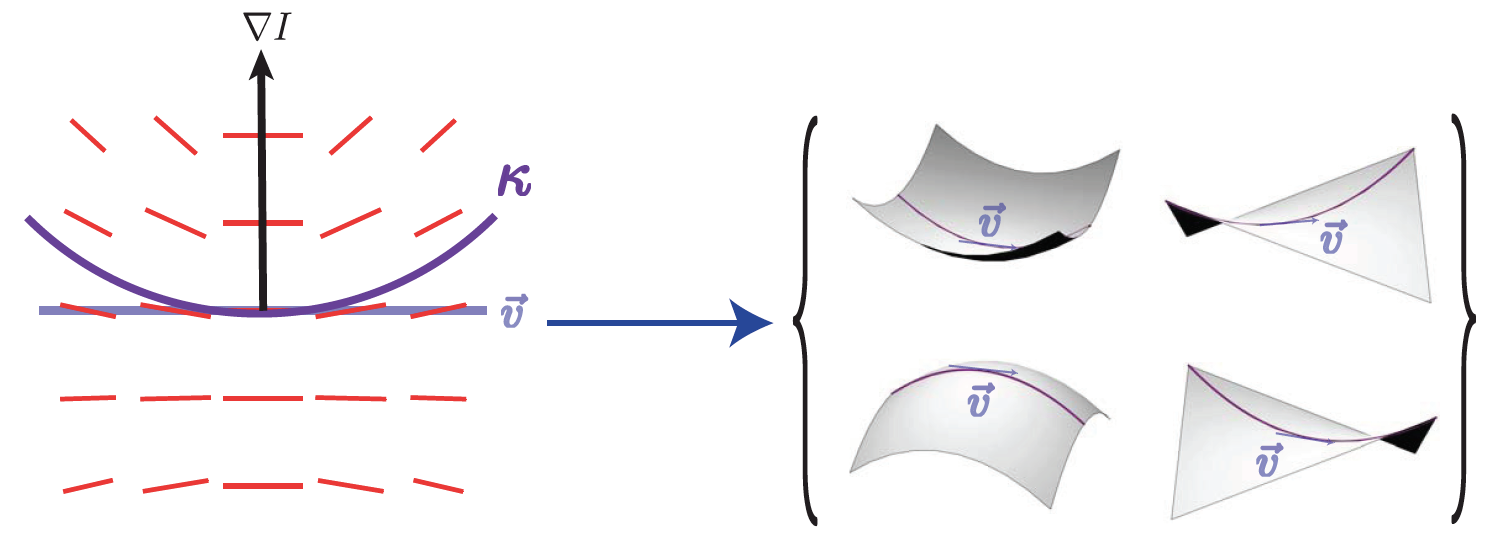} \\
(b) \\
\end{tabular}
\caption{Curvature constraints from patches of shading flow.
(a) Problem setup: a Lambertain object is illuminated from a finite number of point sources. Every patch of the shading flow field supports a family of surfaces (ill-posedness).
(b) Preview of results. A patch of shading flow, the image gradient, and curvature $\kappa$ support a second-order image patch of varying curvature. Figures after \cite{Kunsberg2014}.
\label{fig:first-thread}}
\end{center}
\end{figure}

We begin by considering surfaces rendered through a Lambertian lighting model:
$$I(x, y) = \rho \vec{L} \cdot N(x, y)$$
where the image $I(x,y)$ is formed by the inner product between an (unknown) point source in direction $L$ and the surface normal $N(x,y).$
What can be inferred about a surface without specifying the light source?
The problem setup and a preview of the results are shown in Fig.~\ref{fig:first-thread}. As we show, there are constraints on the surface curvatures. 

We approach this from the perspective of differential geometry. The problem is transformed into the local tangent plane, so we can then use the machinery of covariant derivatives and parallel transport. The key idea is to exploit the fact that intensity remains constant along the isophote. This allows us to represent image derivatives (see Fig.~\ref{fig:steps}) as a function of the surface vector flows.  
 This section is a summary of \cite{Kunsberg2014, kunsberg2014shading, holtmann2018tensors}, which should be consulted for technical details, proofs, and much additional material. 
 
For the analysis, we further assume that the smooth Lambertian surface is a graph of function, with locally constant albedo and Gaussian curvature $K \ne 0$. It can be lit from any finite number of unknown directions and the image is captured through orthogonal projection. The image patch then corresponds to a local surface patch. Using Taylor's theorem, it can be represented: 
$S = \{ x, y, f(x, y) \}$ with $f(x, y) = c_1 x + c_2 y + c_3 x^2 + c_4 x y + c_5 y^2 + c_6 x^3 + c_7 x^2 y + c_8 x y^2 + c_9 y^3.$
The problem then is: {\em Given the shading flow and brightness gradient vector fields, recover constraints on the surface from the local isophote structure.}

In general, to get full information about the curvatures, it is necessary to go to third-order. For our current purposes, however, second-order suffices. Others have also studied this class of surfaces \cite{pentland84, koenderink90Koen}. 

\subsection{The Second-Order Shading Equations}

\begin{figure}[h]	
\begin{center}
\begin{tabular}{c c}
\includegraphics[angle = 90, width=.5 \textwidth]{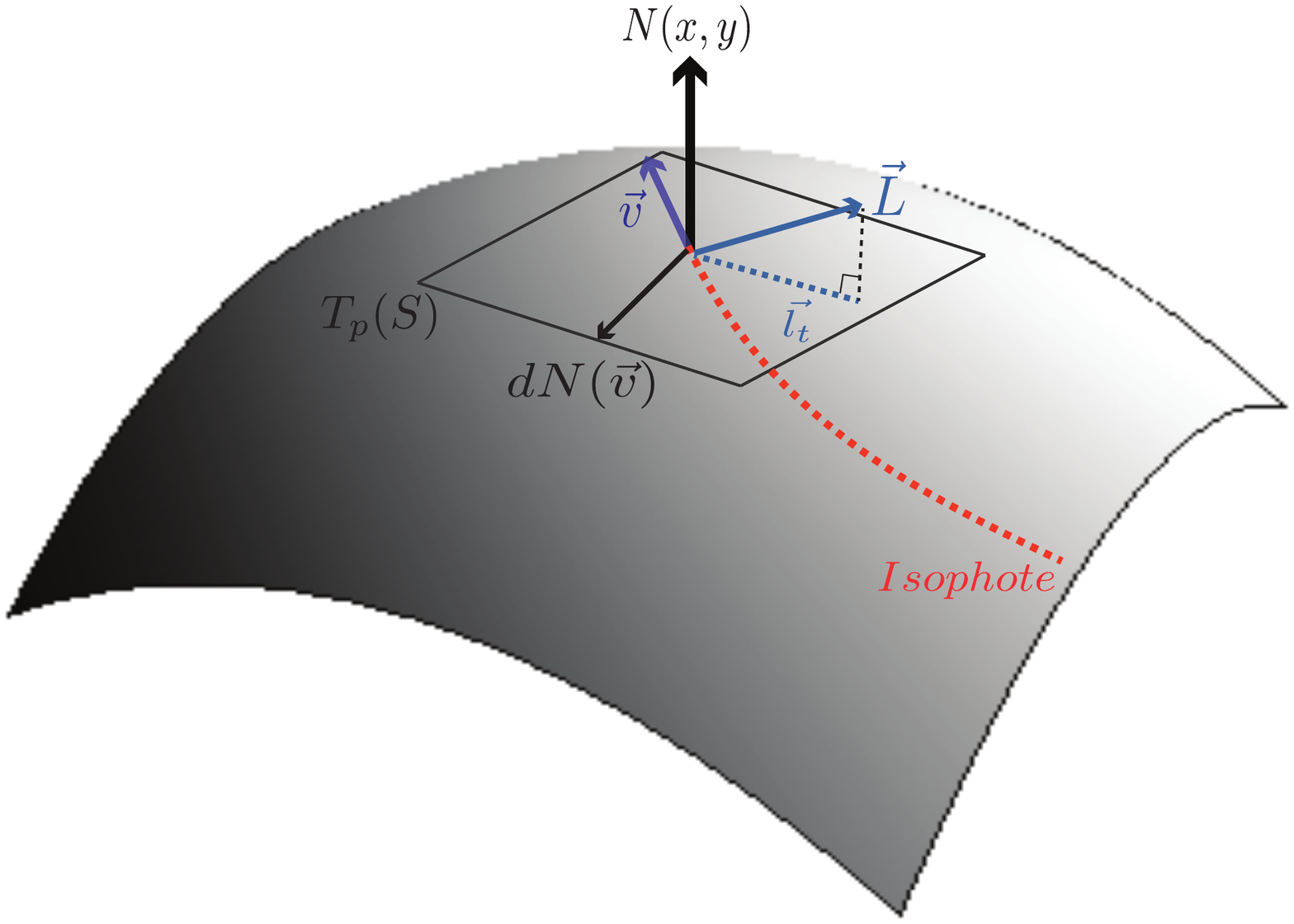} &
\includegraphics[width=.4 \textwidth]{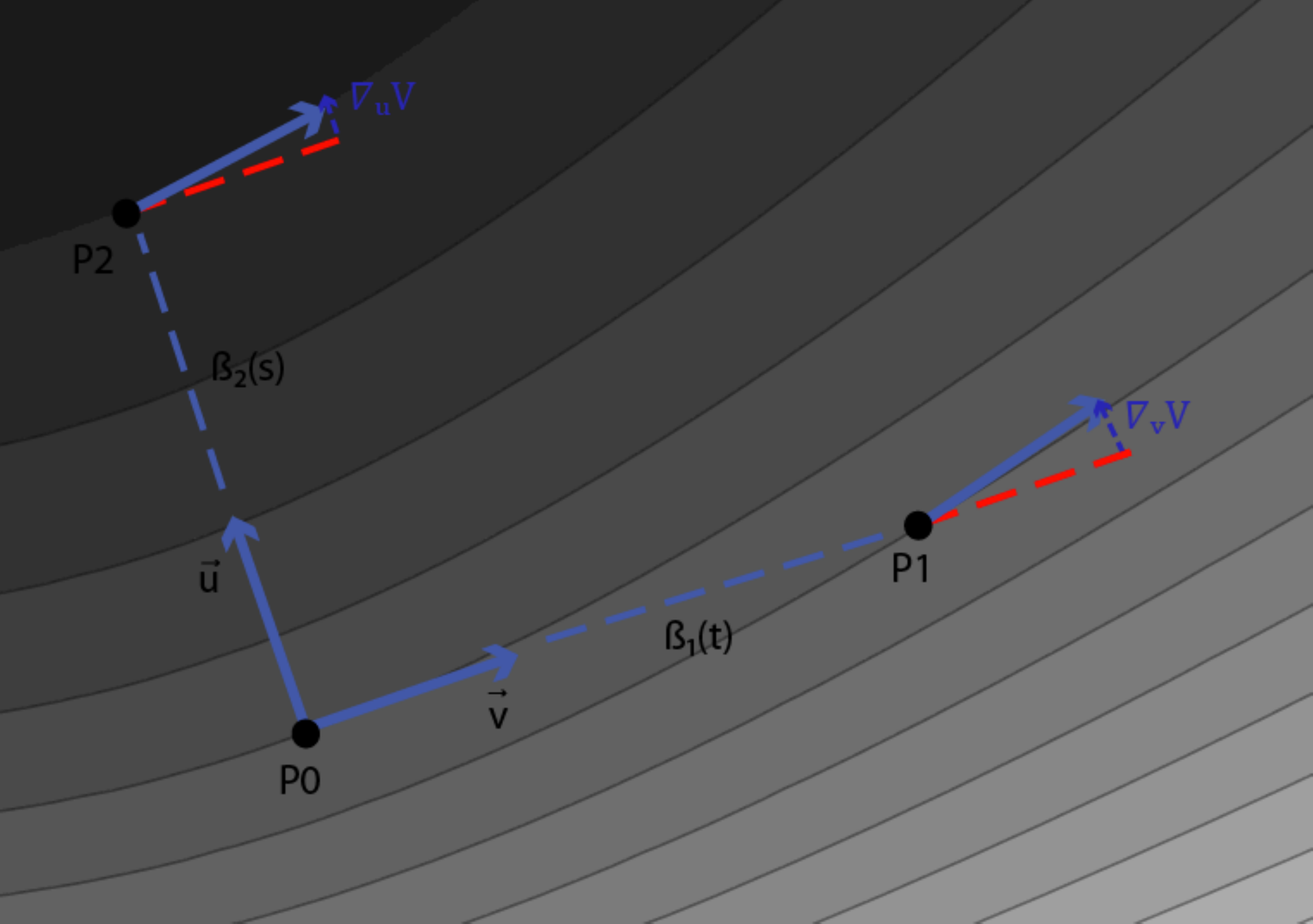} \\
(a) & (b) \\
\end{tabular}
\caption{Differential geometric approach to shape inference. (a) Variables used in our analysis. $N(x, y)$, the unit normal vector field of surface patch $S$; $\vec{L}$, the light source direction;
$\vec{l_t} (p)$, the projection of $L$ onto the tangent plane at point $p$; 
$V(x, y)$, the vector field of isophote directions at each point $(x, y)$, $\vec{v} \in T_p (S)$, the image unit length tangent vector in the direction of the isophote at $p$;
$\vec{u} \in T_p (S)$  the image unit length tangent vector in the direction of the brightness gradient at $p$;
$\hat{w} \in T_p (S)$  the tangent vector in direction $\vec{w}$ of unit length in the image, expressed in the surface tangent basis;
$\vec{u} [ V ]$  the directional derivative of the vector field $V$ in the direction $\vec{u}$;
$\nabla_{\vec{u} } V$;  the covariant derivative of the vector field $V$ in the direction $\vec{u}$. 
(b) Illustration of transports: either along the tangent direction or normal to the tangent direction.}
\label{fig:steps}
\end{center}
\end{figure}

The analysis steps are as follows.  The brightness gradient and isophote are expressed as relationships in the tangent plane between the projected light source and shape operator. Taking the covariant derivative of the projected light source reveals that it is independent of the direction of the light source. Thus it acts like a 'surface property.' We then take the covariant derivative of the isophote condition and separate it into the differentiation on the projected light source and the differentiation on the shape operator. We are finally able to separate the image derivatives into functions of the image and surface properties without reference to the light source direction at all. This gives us a total of three equations equating the second order intensity information (as represented in vector derivative form) directly to surface properties.  

Some notation: 
$V(x, y)$ is the shading flow field, normalized to be unit length in the image -- the corresponding surface tangent vectors have unknown length.   (We denote unit length vectors in the image plane with a vector superscript, such as $\vec{v}$). The corresponding vectors on the surface tangent plane are defined by the image of $\vec{v}$ under the  map composition of the differential $df: \mathbb{R}^2 \rightarrow \mathbb{R}^3$ and the tangent plane basis change $T: \mathbb{R}^3 \rightarrow T_p (S)$.  We will use the hat superscript to denote these surface tangent vectors, e.g. $\hat{v}$. Other notation is in the caption of Fig.~\ref{fig:steps}.

We now state theorem 3.1 from \cite{kunsberg2014shading}, which we call {\em the second-order shading equations}:

\begin{theorem} 
For any point $p$ in the image plane, let $\{ \vec{u} , \vec{v} \}$ be the local image basis defined by the brightness gradient and isophote.  Let $I$ be the intensity, $\nabla I$ be the brightness gradient, $f(x, y)$ be the height function, $H$ be the Hessian, and $dN$ be the shape operator.  Then, the following equations hold regardless of the light source direction:

\begin{equation}
I_{vv} =  - I || dN(\vec{v}) ||^2 +  (\nabla I) \cdot H^{-1} (\vec{v} [H ] \vec{v})
\label{eq:shd-theorem1}
\end{equation}

\begin{equation}
I_{uu} =   - I || dN(\vec{u}) ||^2 - 2 \frac{|| \nabla I ||}{\sqrt{1 + || \nabla f ||^2}} \langle \nabla f, dN(\vec{u}) \rangle +  (\nabla I) \cdot  H^{-1} \cdot (\vec{u} [H ] \vec{u})
\label{eq:shd-theorem2}
\end{equation}
\end{theorem}

\begin{equation}
I_{uv} =  - I \langle dN(\vec{v}), dN(\vec{u}) \rangle - \frac{|| \nabla I ||}{\sqrt{1 + || \nabla f ||^2}} \langle \nabla f, dN(\vec{v}) \rangle +   (\nabla I) \cdot H^{-1} \cdot (\vec{u} [H ] \vec{v})
\label{eq:shd-theorem3}
\end{equation}

Note that there is no dependence on the light source, except through measurable image properties and their derivatives. Furthermore, these equations directly restrict the derivatives of local surface patch. To see the interplay, it is helpful to write the first equation as 
\begin{equation}
    \frac{\nabla I \cdot \kappa}{I} = ||dN (\vec{v}||^2
\end{equation}
which was illustrated in Fig.~\ref{fig:first-thread}. Note that the image properties are on the left side, and surface curvature properties are on the right. Since the shape operator appears inside $||\cdot||^2$, any rotation will suffice. This illustrates the ambiguity and directly allows for saddles. In general, even for this restricted patch analysis, there are 5 variables (tangent plane orientation and surface curvatures) but only 3 equations. Taking further derivatives reveals more possible variation \cite{holtmann2018tensors}. Furthermore, it might be thought that the image gradient constrains the direction of the light source (e.g., \cite{koenderink04}), but the above equations show that this is not necessarily the case unless priors are introduced. 

The insight that we got from these equations is different. Since the form of the above equations simplifies substantially at critical points, or places where the image gradient is 0, this became our focus. We now turn to one such case, to begin the transition into a different way of thinking about shading inferences.

\section{Transition}
\label{sec:transition}

\subsection{Ridges}

Ridges are very salient regions of an image; they can sometimes be isolated via a Laplacian filter. While intuitively clear, there remains disagreement about how to define ridges. For now we shall adopt a very simple working definition \cite{eberly1994ridges, lindeberg1998edge, haralick1983ridges}, and will use the term ``ridge" as an image contour of connected points all of which are either local minima or maxima of intensity and, for now, will assume $K = 0$.  Along these contours, we get a simplification of the shading equations. (These non-generic notions will be relaxed significantly in the second thread.)

Let $\vec{w} \in T_p S$ denote the (unknown) nonzero principal direction; it corresponds to the major axis for a (locally cylindrical) Taylor approximation.  Specifying a Frenet basis for this ridge image contour, we can express the surface derivatives in this $\{\vec{u}, \vec{v} \}$ basis. Let $\vec{l_t}$ be expressed as  $\{l_1, l_2\}$ in this basis. Although $\vec{w}$ is unknown, at highly foreshortened tangent planes (as is the case on the ridges in this example), tangent vectors project to either $\{ \vec{u}, \vec{v} \}$, up to the resolution of the visual system.  Without loss of generality, suppose $\vec{w} \approx \vec{u}$.  Then, the shading equations reduce to:

\begin{subequations}
\begin{align}
I_{vv} & =  - (I) || dN(\vec{v}) ||^2  + \frac{l_2 f_{vvu}}{\sqrt{1 + || \nabla f ||^2}}\\ 
I_{uu} & =    \frac{l_2 f_{uuu}}{\sqrt{1 + || \nabla f ||^2}} \\
I_{uv} & =  \frac{l_2 f_{vuu}}{\sqrt{1 + || \nabla f ||^2}} 
\end{align}
\end{subequations}

\begin{figure}[!ht]	
\begin{center}
\begin{tabular}{c c c}
\includegraphics[width=.3 \textwidth, keepaspectratio]{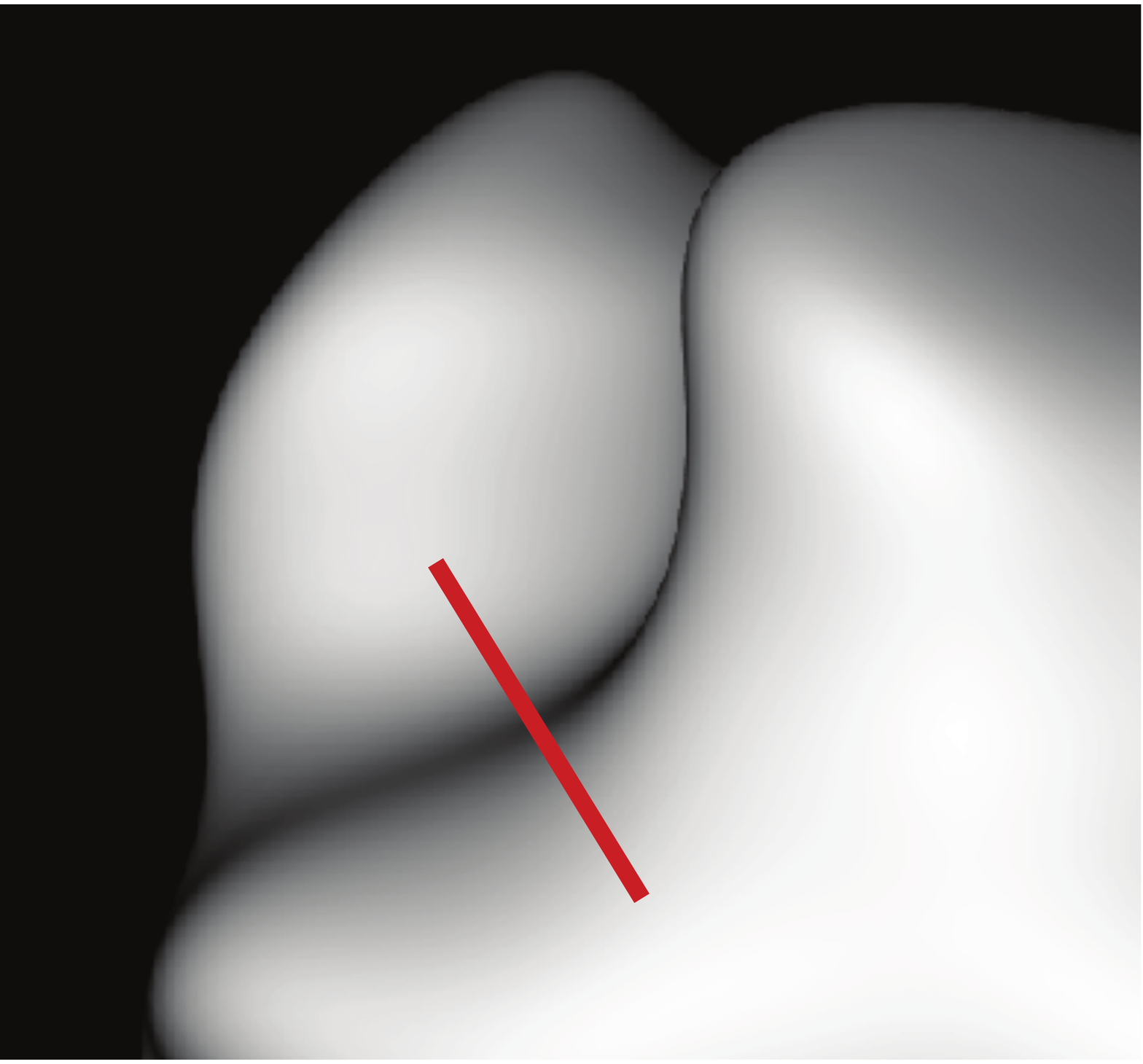}  &
\includegraphics[width=.3 \textwidth, keepaspectratio]{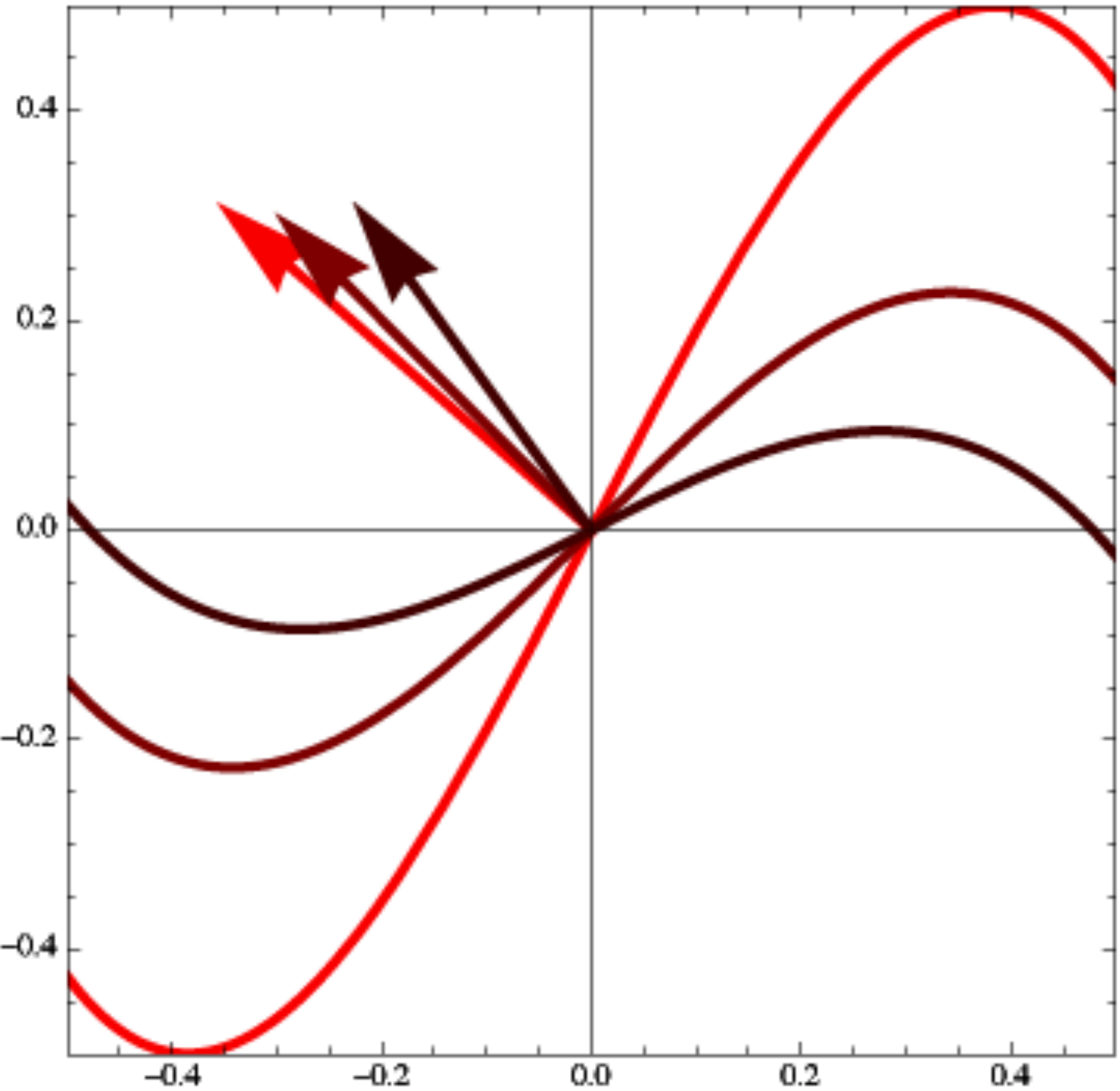}  &
\includegraphics[width=.3\textwidth, keepaspectratio]{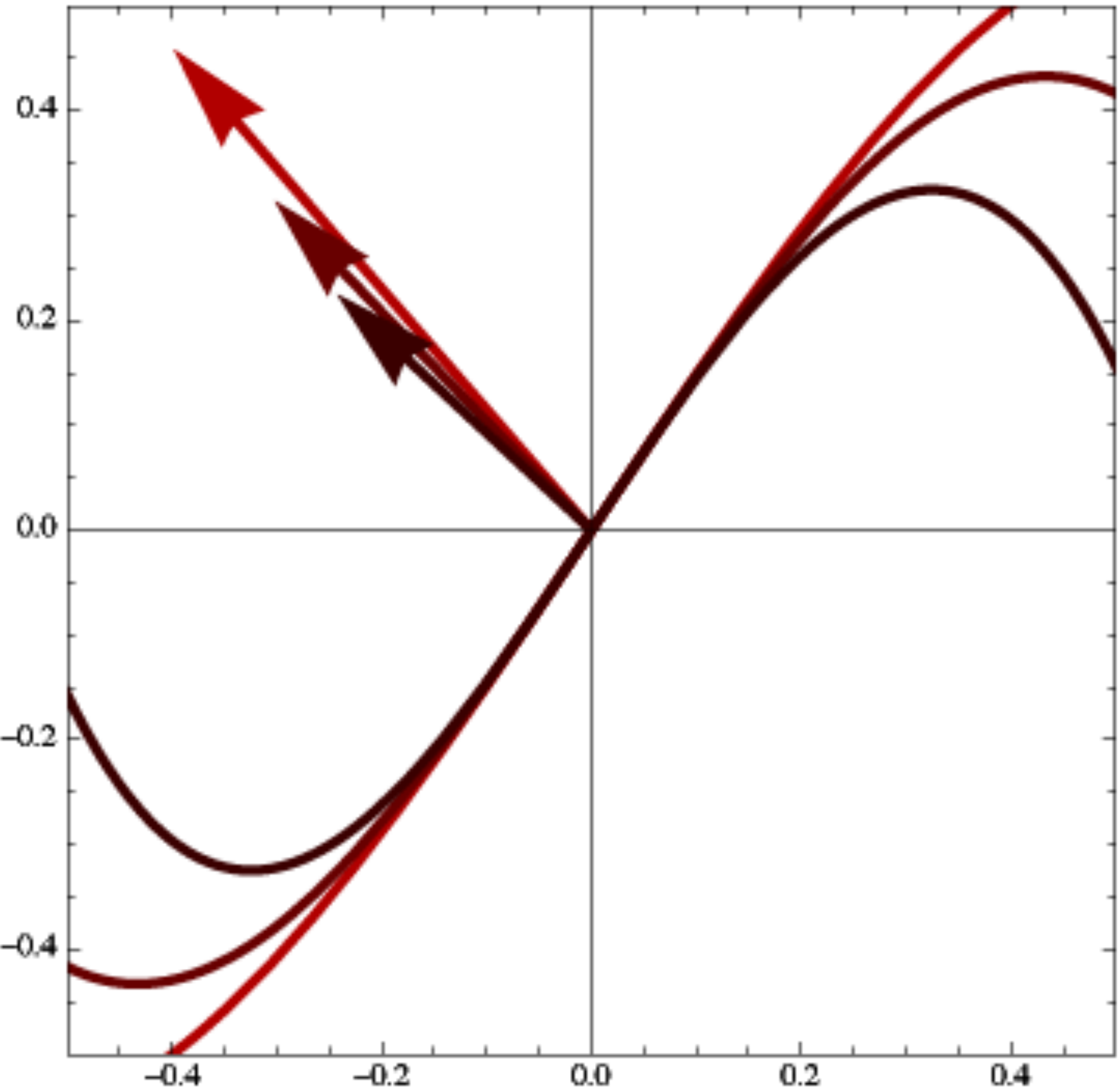}  \\
(a) & (b) & (c) \\
\end{tabular}
\caption{Special structure along ridges.
(a) A shaded surface.  The red line defines a normal plane that cuts across the riged. (b) The curves represent various possible cross sections resulting in the same $I_{uu}$ value, and the arrows represent the necessary light source.  Note the co-variation between the tangent plane and the light source. (c) Cross sections and associated light sources with the tangent plane fixed; the projected light source changes. Not the similarity in form of the different possible cross sections.}
\label{fig:ridge_cross_section}
\end{center}
\end{figure}

Note that both $I_{uu}$ and $I_{uv}$ are proportional to third derivatives of the surface, with the coefficient given by the foreshortening of the tangent plane and light source.  That is, ambiguity in several Taylor coefficients is only a matter of scale.   Perhaps the most relevant of these three equations is the first, which shows how weighted versions of $|| dN(\vec{v}) ||$ can be traded off with $f_{vvu}$ to keep image properties the same. Geometrically this sparked our intuition (Fig.~\ref{fig:ridge_cross_section}).

\subsection{Stable Neighborhoods}

\begin{figure}[h]	
\begin{center}
\includegraphics[width=.6\textwidth]{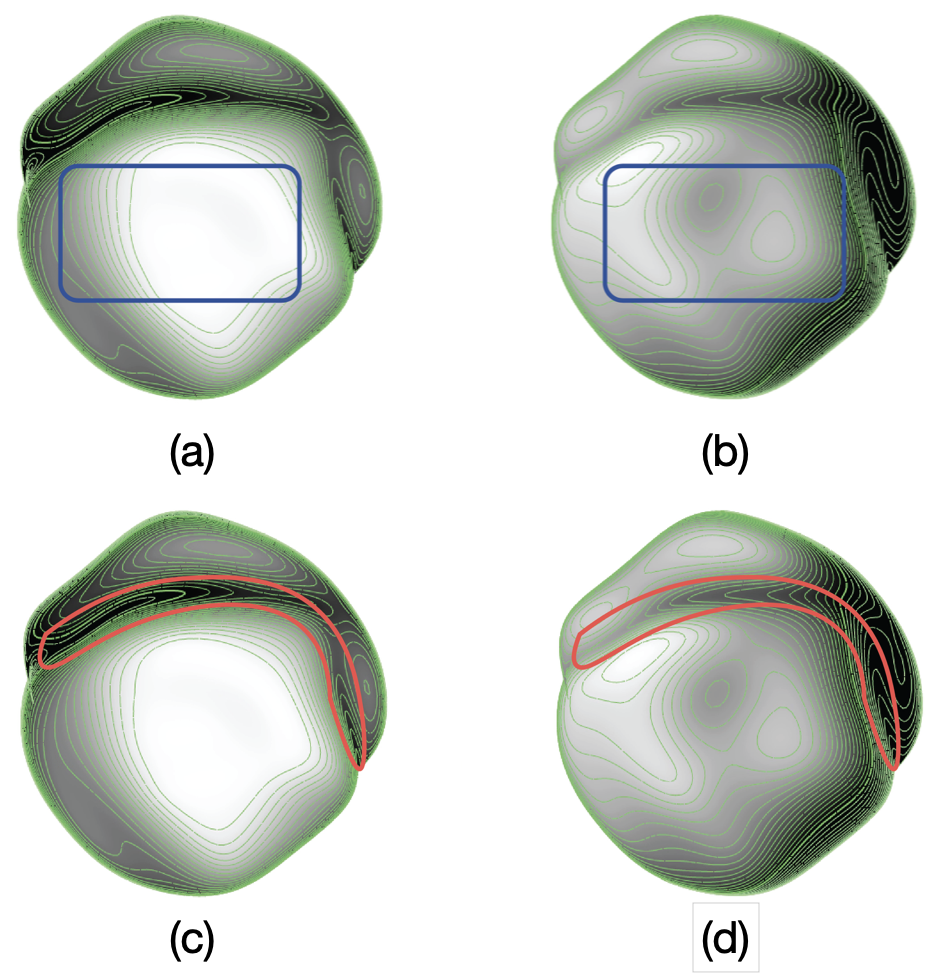} 
\caption{Stable isophotes and illumination on a random, smooth Lambertian shape. (a, b) (blue box) Note the change in the isophote structure with a change in illumination. (c,d) (red sausage) Note the similarity of isophotes across lighting. A ridge line cuts through them in the normal (i.e. gradient) direction. }
\label{fig:stable-isophotes}
\end{center}
\end{figure}

In the introduction to Thread 1, we argued that isophotes and the shading flow provided the input on which shape inferences were grounded. By focusing on Lambertian reflectance models, we avoided critiques of this position, namely  that the isophotes change drastically with rendering changes, and even with illumination changes (e.g. \cite{doi:10.1068/i0645}). This critique is substantial; see Fig.~\ref{fig:stable-isophotes}(a,b). The isophotes within the blue box change drastically with a change in illumination. If the orientations vary so much, how does the 3D perception remain consistent? Perhaps the isophotes do not change this much everywhere. In particular, the analysis of the second-order shading equations suggested where one might look. While still ill-posed, certain neighborhoods -- those involving ridge singularities -- might be special. Importantly, there was a kind of "form" to the shape in these regions, even though the exact shape itself remained ill-posed.

A closer look at the isophotes in an open neighborhood around the ridge reveals something remarkable: they take on a stereotyped form. If one could draw a contour along a general ridge like these, then this contour would cut directly across the nested isophotes. This observation is the key to the transition, and the key to relating shape inferences from shading to shape inferences from contours. One might think of {\em the contour as a concentration of shading.} This is precisely what we do in the second thread.

\section{Thread II: From Shading Flow to Critical Contours}
\label{sec:thread2}

A masterful sketch can give rise to a vivid 3D shape percept. The information is a set of lines and their arrangement, including the corresponding white space.  How might a brain 'reconstruct' from this impoverished information?  Clearly it is impossible to find \emph{the} veridical percept; however, as we have all experienced,  qualitative judgments should still be possible. Is this also the case for shape-from-shading, and shape-from-texture? Are these all versions of the same problem?  We now argue constructively that this is indeed the case, by developing a scheme for doing it. The development follows several steps. First, we argue that the perception of shape is qualitative, not quantitative, a point that has been well understood in visual psychophyics for decades. This suggests that we should not be seeking to solve the shape-from-shading equations, but should look for qualitative (that is, topological) solutions instead. The Morse-Smale complex emerges as the natural substrate. Clearly this cannot take us all the way to an exact shape; instead we obtain a kind of scaffold on which the shape can be readily built. Technically, we show that the contours -- what we call critical contours -- are a type of shading concentration, and are also part of the Morse-Smale complex. Importantly, this holds for a wide range of rendering and lighting functions. And to connect back to the ridge example, it describes precisely what is going on in the red sausages in Fig.~\ref{fig:stable-isophotes}. The material in this section summarizes that in \cite{Kunsberg18, Kunsberg2018Focus, kunsberg2020boundaries}. 

\subsection{Shape perception is qualitative}
\label{sec:qual}

\begin{figure}
\begin{center}
\includegraphics[width = 0.8 \linewidth]{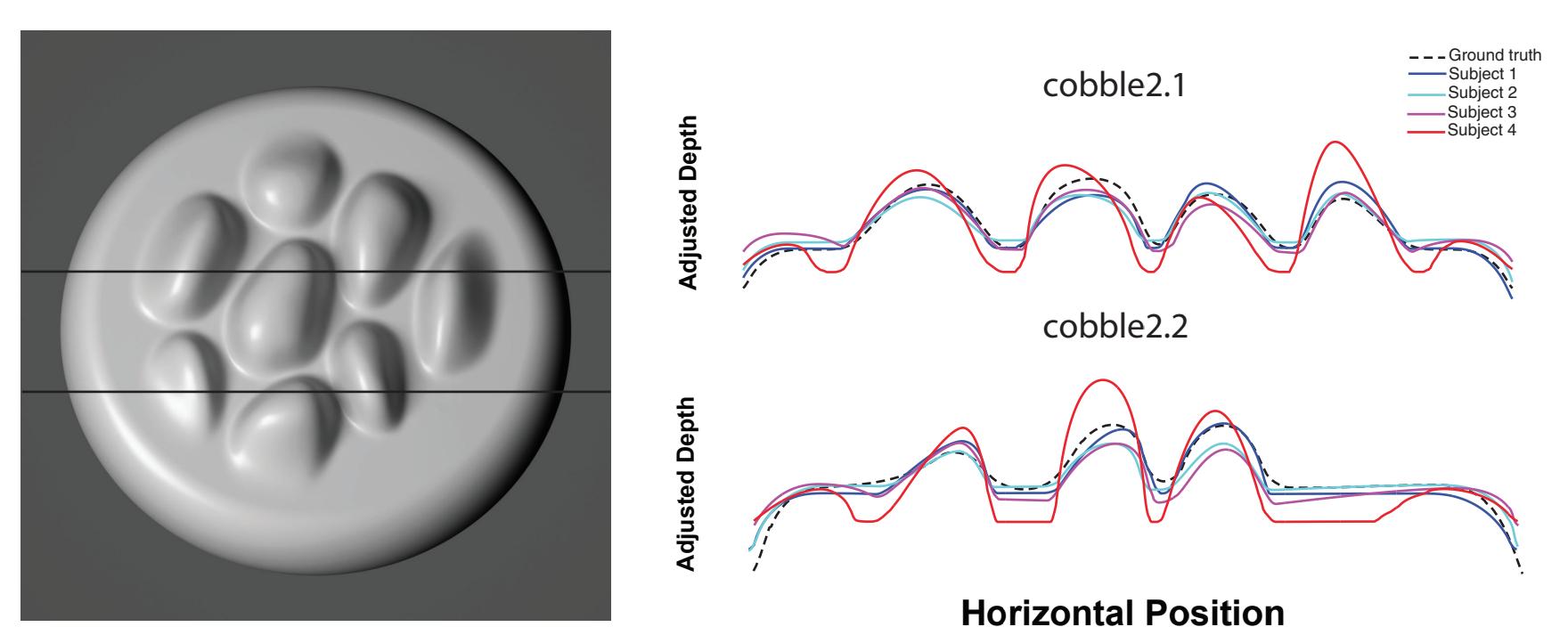}  \\
\caption{Bump perception is qualitative (left) An image showing a number of bumps on a surface. Superimposed are two scanlines, along which subjects estimated the height. The results (right) show that each subject perceived a slightly different profile, although there is substantial agreement in the general outline: we say that the heights and the boundaries are in qualitative, but not quantitative agreement.  Figure from \cite{ToddVSS17} \label{fig:todd}}
\end{center}
\end{figure}

Although we have the sense that, when looking at a shaded object we see it {\em exactly} as it is in the world, this is an illusion. There is a large literature on the psychophysical evaluation of perceived shape, and it has characterized the many differences both within and between subjects when viewing a shape. The idea is to have subjects estimate either the surface normal (or an equivalent gauge figure) or the relative depth of a shape, at many different positions, and then integrate these into a representation of the internal shape. An example is shown in Fig.~\ref{fig:todd} from \cite{ToddVSS17}. While all  subjects were in agreement about the qualitative nature of the bumps, there was substantial quantitative variation across them. 

Humans do not perceive veridical depths; the perceived heights of the `stones' in Fig.~\ref{fig:todd} are not only literally incorrect but differ across individuals. Nevertheless, the perceived shapes agree in the localization of the 'bumps' on the surface, i.e., the edges of the `stones' are accurately perceived. Such qualitative variability yet agreement must be part of any shape inference scheme intended to model human perception. Similiar results hold for a variety of different rendering functions and many different studies show similar results \cite{doi:10.1167/12.1.12, MAMASSIAN19962351,Mingolla1986, doi:10.1068/i0645, doi:10.1167/15.2.24, CHRISTOU19971441, doi:10.1167/15.2.24, Seyama19983805,Curran19961399, doi:10.1068/p5807, doi:10.1068/p251009}. 

\begin{remark}
We now specify a shape representation -- critical contours -- that captures the above phenomena. It involves three steps. First, we specify how contours can be viewed as concentrated shading. Second, we observe that these contours are part of the gradient flow. Third, we cite the main theorem in this thread, that critical contours are part of the Morse-Smale complex generically; that is, for different rendering functions. 
\end{remark}

\subsection{Critical Contours as Concentrated Shading}

\begin{figure}[ht!]
\begin{center}
\begin{tabular}{c}
\includegraphics[width = 0.65 \linewidth]{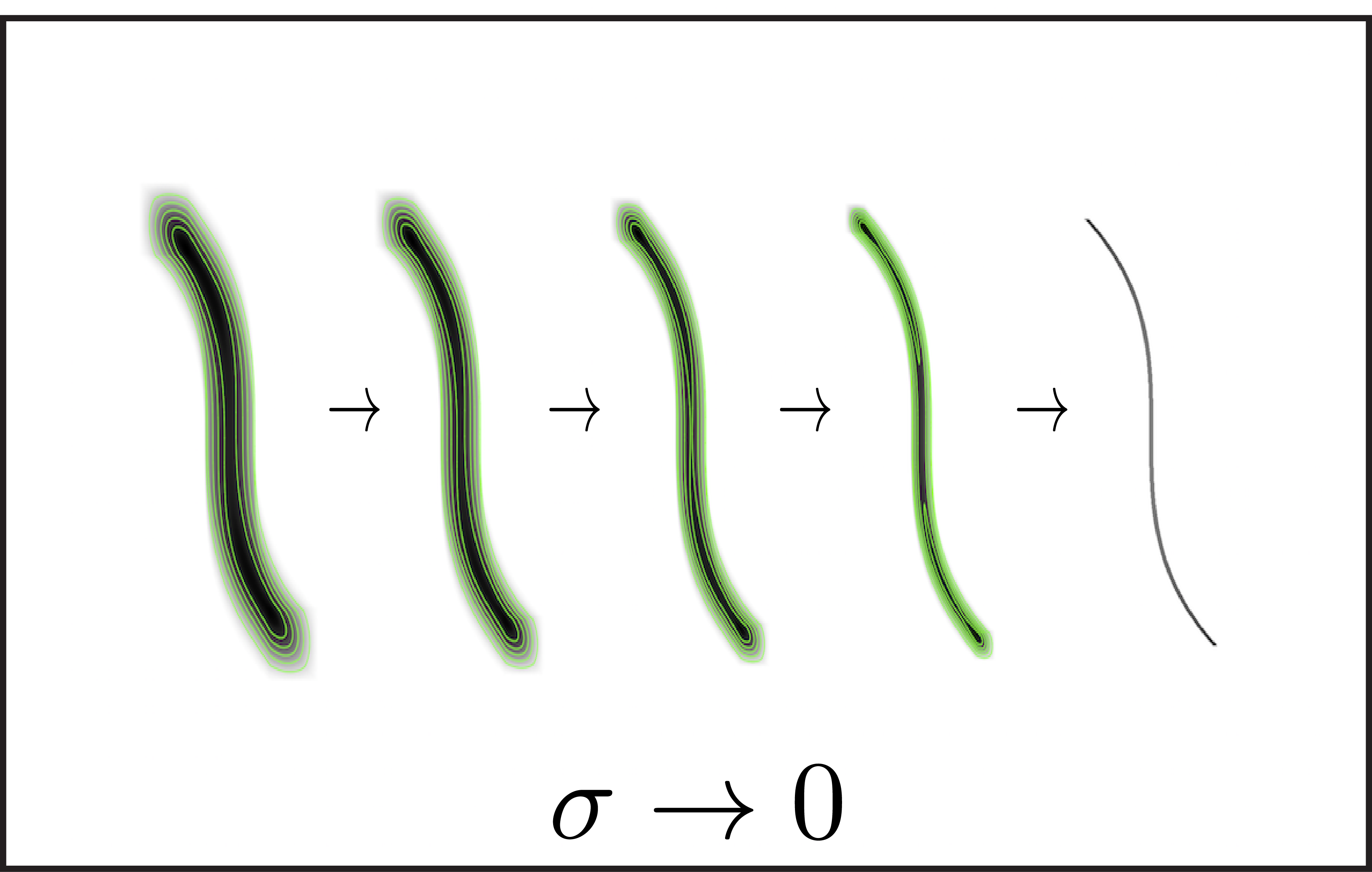} \\
(a) 
\end{tabular}
\begin{tabular}{c c}
\includegraphics[trim= 1cm 0cm 0cm 0cm, clip=true, width = 0.3 \linewidth]{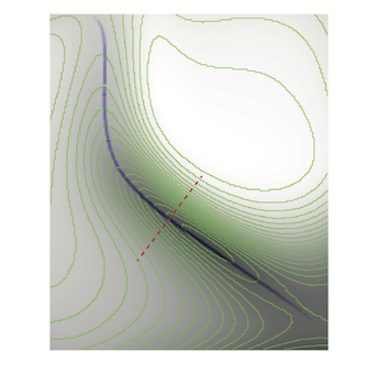} &
\includegraphics[width = 0.33 \linewidth]{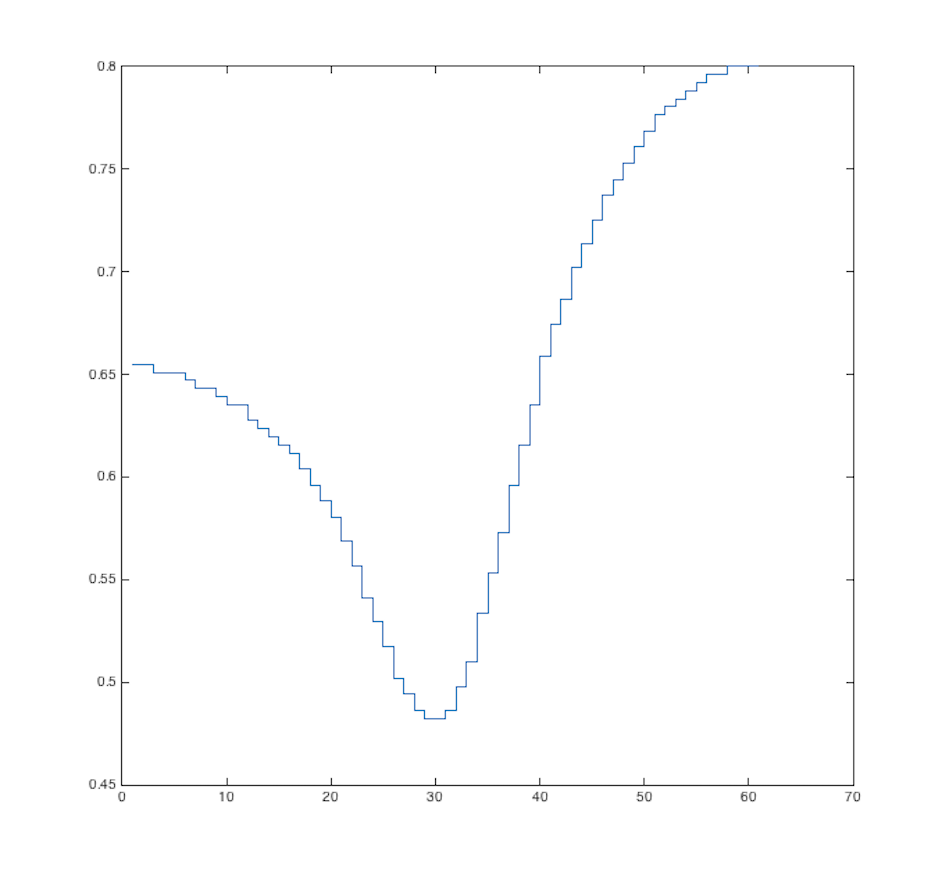} \\
(b) & (c) \\
\end{tabular}
\caption{The meaning of certain contours as a limit of shaded images.
(a) Concentrating the blur along a contour. Note the organization of isophotes within the blurred regions. $\sigma$ denotes the degree of blur in the sequence of images $I^\sigma_\alpha$. (b) A portion of a shaded figure, with isophotes (in green). The blue contour outlines the 'ridge' section. Note the similarity of the two isophote patterns near the contour: on either side of the contour the direction of the level curves rotate to be nearly tangential to the contour.  Also, note how the contour (blue) is nearly a gradient flow of the shaded image.  (c) The intensity values along the dotted red line. Note the steep local shading minimum across the contour. The blue curve is a portion of a critical contour.}
\label{fig:transvers}
\end{center}
\end{figure}

The bumps in the last figure were outlined by a dark region; this surrounded a lighter central region and a lighter base from which it extended. Similarly, for the ridge example, the top was light, the slope was dark, and the bottom was light again. If an artist were to draw the ridge, she would likely draw a contour somewhere in this dark region. We formalize this as a concentration of shading phenomenon (Fig.~\ref{fig:transvers}).  

We consider a line drawing image as a collection of these 1D contours.  Consider one contour, $\alpha(t)$, arclength parametrized, with bounded planar curvature. Let each value along $\alpha (t)$ be the scalar intensity value $I(\alpha (t))$; it should be 0 at the endpoints.  

To relate the curve to a sequence of shaded images, we first make its trace into an image and then blur it with a gaussian. This yields a sequence of (shaded) images $\{ I_{\alpha}^\sigma \}, \sigma \rightarrow 0$ on  $\Omega_\alpha$.  Each shaded image $ I_{\alpha}^\sigma$ is then a convolution with appropriate Gaussian functions $G(\sigma)$ of $I_\alpha$ with successively smaller standard deviation. For details, see \cite{Kunsberg18}. 

To parameterize the shading in a local neighborhood $U_\alpha(t)$ around $\alpha(t)$, let $u(t) = \alpha'(t)$ and define $w(t)$ so that $w(t) \cdot u(t)= 0$.  At the point $p$, $\{u, w \}$ is an orthonormal basis.

We now define a \emph{critical contour}, which will turn out to be the (nearly) invariant visual pattern across renderings. As in Fig.~\ref{fig:transvers}, it has slowly varying intensity along it and "high walls" across it:

\begin{definition} 
\label{defn_cc}
A $K$-Critical Contour $(\alpha(t), I(x, y), M, K)$ is a curve $\alpha(t)$ on an image $I(x, y)$ such that the following conditions hold for all $t$:
\begin{enumerate}
\item $| I_{ww} (\alpha(t))| > K$ for every $t \in [0, 1]$.
\item $| I_{uu} (\alpha(t))|  > K $ for $t = \{0, 1\}$.
\item $| DI |, |I_{uw}| < M$ for every $t \in [0, 1]$.
\end{enumerate}

\end{definition}

As $K \rightarrow \infty$, K-critical contours converge pointwise to the ideal contour.

Critical contours are part of the gradient flow, which opens the door to a remarkable connection to topology. We turn to this next.

\subsection{The Morse-Smale Complex}

The Morse-Smale complex characterizes the topology of surfaces in terms of the singularities of a scalar function on that surface. We will work with two different classes, either the image intensity function or the slant function $\sigma(x, y): \mathbb{R}^2 \rightarrow \mathbb{R}$ on the surface. The requirement is that they be \emph{Morse functions}: all  critical points must be non-degenerate (i.e., the Hessian at those points is non-singular) and no two critical points have the same function value. This introduction is from \cite{kunsberg2020boundaries}; for additional material, see \cite{milnor2016morse, Gyulassy08, Biasotti:2008:DSG:1391729.1391731, matsumoto2002introduction}.

\begin{figure}[ht!]
\begin{center}
\includegraphics[width=0.6 \linewidth]{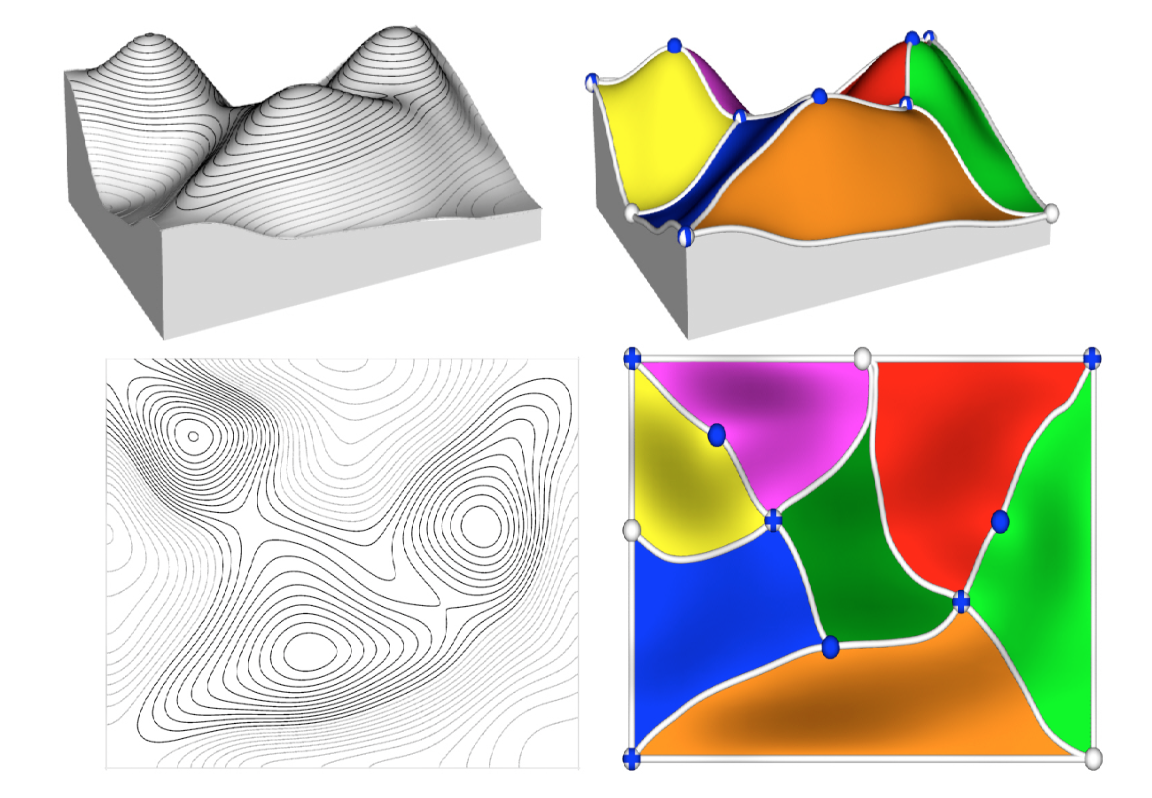}
\caption{Illustration of the Morse-Smale complex as a topological invariant of surfaces. (left) A scalar function and its level sets, shown in two views. If this function were image intensity, the level sets would be isophotes. (right) The colored diagram is the MS-complex, superimposed on these figures.  Colored  regions  represent 2-cells  of  the  MS  complex. White curves represent 1-cells (contours) of the MS complex.  Maxima, saddles, and minima are represented by solid blue points,  crosses,  and solid  white  points,  respectively.  Each 2-cell  is  combinatorially  a quadrilateral,  but  sometimes two  saddles  can  be  identified  together creating  a  loop.  This illustrates  how  a  set  of 2D contours can represent a skeleton of a 3D surface, up to monotonic transformations on each region. Figure courtesy of Attila Guylassy. \label{fig:MS_example}}
\end{center}
\end{figure}

For a smooth Morse surface, the \emph{gradient} 
$\nabla \sigma = \left( \partial f / \partial x, \partial f / \partial y \right)$
exists at every point.  A point $p$ is a \emph{critical point} when $\nabla \sigma (p) = 0$.  This gradient field gives a direction at every point in the image, except for the critical points, a set of measure zero. Following the gradient vector field will trace out an \emph{integral line}. These integral lines must end at critical points, where the gradient direction is undefined.  Thus, one can define an \emph{origin} and \emph{destination} critical point for each integral line.  

The type of each critical point is defined by its \emph{index}: the number of negative eigenvalues of the Hessian at that point.  For scalar functions on $\mathbb{R}^2$, there are only three types: a maximum (with index 2), a minimum (with index 0) and a saddle point (with index 1).  

There are two types of integral lines, depending on the difference in index of the critical points that it connects.  If the difference is one, the integral line is called a \emph{1-cell}; it must connect a saddle with either a maximum or a minimum.   For example, a \emph{saddle-maxima 1-cell} connects a saddle and a maximum.  The set of 1-cells will naturally segment the scalar field into different regions, called \emph{2-cells}.   In addition, the scalar values on the 1-cells govern the values on the 2-cells.  See Fig \ref{fig:MS_example} for insight.

Further, for each critical point,  its \emph{ascending manifold} is defined as the union of integral lines having that critical point as a common origin.  Similarly, its \emph{descending manifold} is the union of integral lines with that critical point as a common destination.   


 For two critical points $p$ and $q$, with the index of $p$ one greater than the index of $q$,  consider the intersection of the descending manifold of $p$ with the ascending manifold of $q$.  This intersection will be either a 1D manifold (a curve called a 1-cell or watershed) or the empty set.   For two critical points $r$ and $s$, with the index of $r$ two greater than the index of $s$, the intersection of the descending manifold of $r$ with the ascending manifold of $s$ will either be a 2D manifold (a region called a 2-cell) or the empty set.  Thus, the intersection of all ascending manifolds with all descending manifolds partition the manifold $\mathbb{M}$ into 2D regions surrounded by 1D curves with intersections at the critical points.

The Morse-Smale complex is the combinatorial structure (and the corresponding attaching maps) defined by the critical points, 1-cells and 2-cells.  It is a structure that relates a set of contours (the 1-cells) to a qualitative function representation.  With knowledge only of the slant function at the critical points and 1-cells, one could reconstruct the 2-cells (and thus the entire function) relatively accurately.  For some insight, see \cite{Giorgis15, Weinkauf10}.  

We now show how the critical contours are 1-cells, and how they can be used as a shape representation.

\subsection{The Main Theorem}

To relate critical contours to the Morse-Smale complex, we first need to specify an image formation model. For simplicity, assume orthogonal projection and align the 3D coordinate axis $(x, y, z)$ so that $(x, y)$ parametrize the image plane and $z$ is the view direction.  The image is formed by applying a rendering function $F: \mathbb{S}^2 \rightarrow \mathbb{R}$ in $C^2$ to the normal field $N(x, y) = (n_1, n_2, n_3)$ of a smooth surface $S(x,y)$.  That is, $I(x, y) = F(N(x, y))$. Many familiar cues have this structure, e.g. Lambertian shading, the spatial frequency cue of (blurred) isotropic texture, specular shading, and so on. Our goal is to seek image contours that are present regardless of the choice of $F$.

Two natural constraints are necessary to define an admissible class of rendering functions. First, a bounded variation condition ensures that arbitrarily large changes in the image cannot be due to the rendering function alone but must also involve a change in the normal field.  Without this, it would be impossible to differentiate material changes (such as a painting) from gradients due to shading.  Second, a concave condition ensures that, if the unit sphere were imaged with a rendering function, there would only be one highlight (point  of  maximum  brightness). Finally, we also require that the surface be in general position with respect to the rendering function. For a technical version of these conditions, see \cite{Kunsberg18}.

The idea is to show that, if there is a critical contour in an image of a surface from one rendering function, then there is also one from any other rendering function in the admissable class.

\begin{figure}[h!]
\begin{center}
\begin{tabular}{c c}
\includegraphics[width=.3\linewidth]{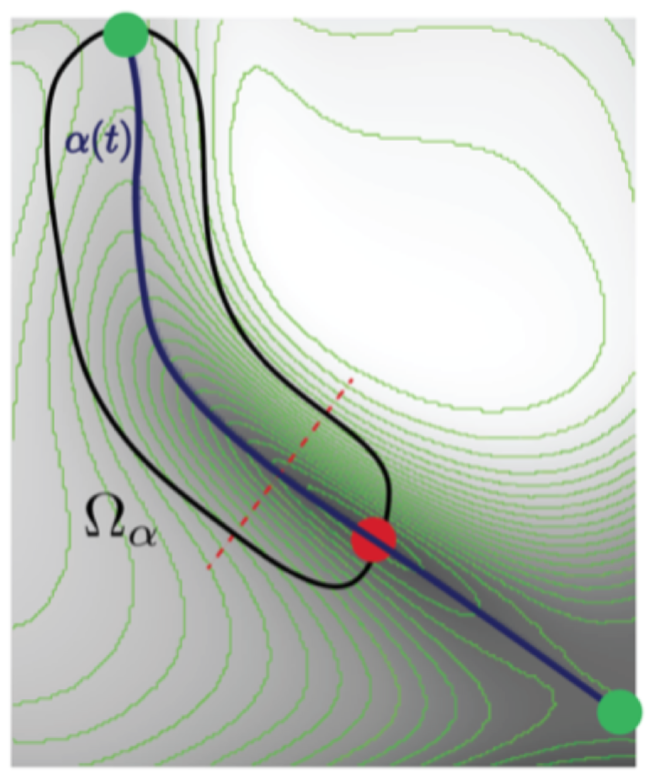} &
\includegraphics[width=.5\linewidth]{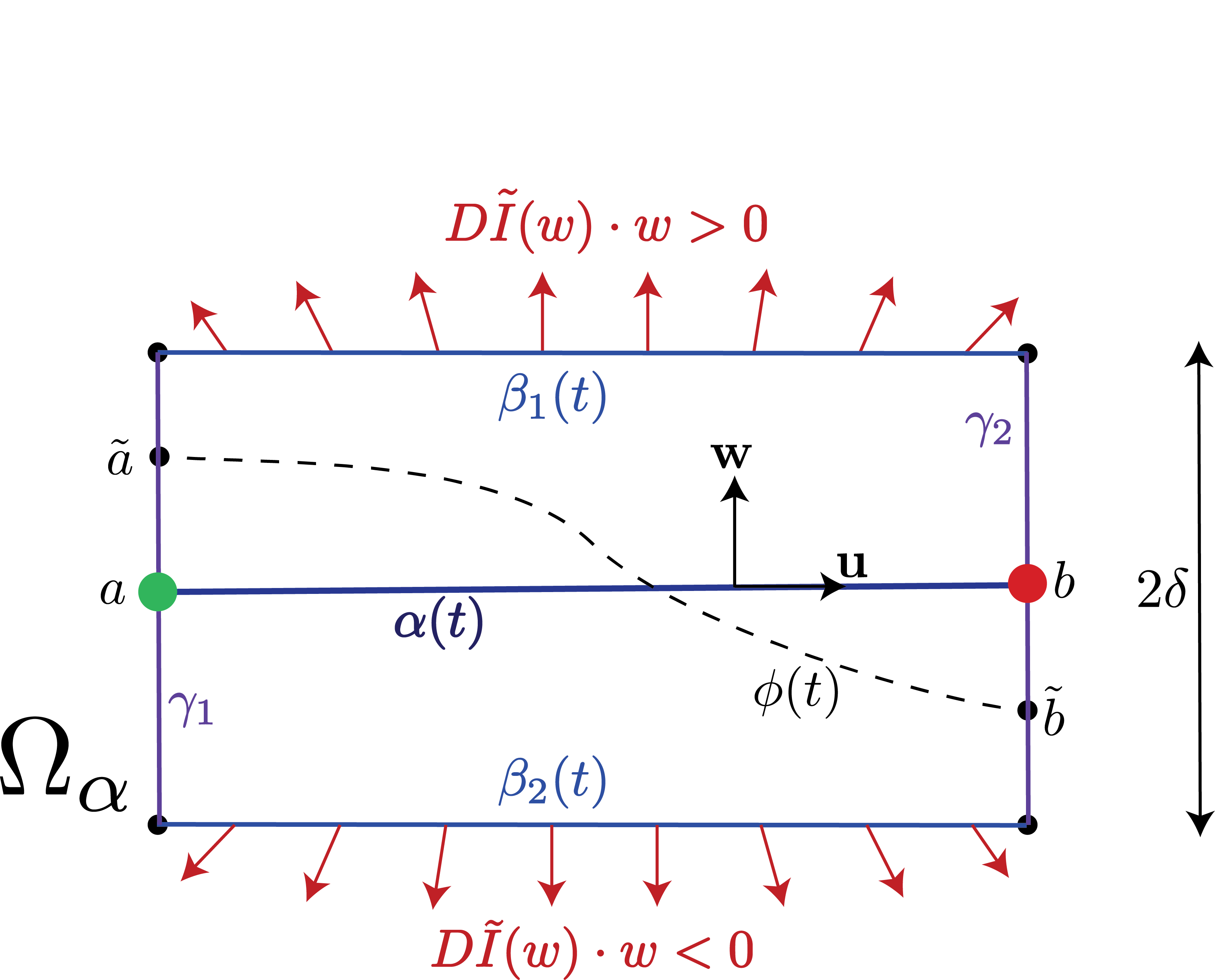}\\
(a)  & (b) \\
\end{tabular}
\caption{\label{fig:theorem2} Illustration of theorem.
(a) The critical contour (blue curve) $\alpha(t)$ is a distinguished curve passing through two singularities (a saddle and a minimum). It is also a gradient flow integral curve. 
(b) The red sausage in (a) becomes a box, with $\alpha(t)$ passing through the center. According to the theorem, there is another critical contour $\phi(t)$ that is nearby. Notice how the gradient points away from the critical contours. }
\end{center}
\end{figure}

\begin{theorem}{1}
\label{main_theorem}
Let $F, \tilde{F}$ be any two rendering functions in the admissible cue class.  Applying these rendering functions to a generic surface $S$, we obtain two corresponding images $I(x, y), \tilde{I}(x, y)$.  For any $\delta > 0$, there exists a $K \in \mathbb{N}$ such that the surface region corresponding to a 
$\delta-$neighborhood of  a $K$-critical contour in $I$ contains a Morse Smale 1-cell for image $\tilde{I}$.
\end{theorem}

The proof is technical, and may be found in \cite{Kunsberg18}. We here illustrate its central aspects (Fig.~\ref{fig:theorem2}). Critical contours pass through critical points and are integral curves though the gradient field, with the additional condition of "steep sides," i.e. the gradient is steep and points away on both sides. Now, consider a critical contour $\alpha(t)$ in an image I, and another critical contour $\phi(t)$ in a different image $\tilde{I}$ from another rendering function. Then there are critical points $\tilde{a}$ and $\tilde{b}$ near the critical points $a$ and $b$ such that, as the sides become infinitely steep (the limit $\sigma \rightarrow 0$) the critical points and the critical contours converge.

It could be the case, for example, that the images $I$ and $\tilde{I}$ correspond to the same surface but illuminated differently, as in Fig.~\ref{fig:stable-isophotes}. A deeper version is to consider the rendering function more directly as a function on the surface. In particular, \emph{slant} is the polar angle between the surface normal and the view direction, and it can be considered as a scalar function on the image domain.  

While in general the slant function is unknown when the surface is unknown, the following corollary establishes a one-to-one relationship between an image-computable function and a surface-computable function. This is Corollary 10 in \cite{Kunsberg18}.

\begin{corollary}
For K sufficiently  large,  a K-critical  contour $\alpha(t)$ in  image $I$ of  surface $S$ aligns with an MS1-cell of the slant of the surface normal field of $S$.
\end{corollary}

\begin{figure}[ht!]
\centering
\includegraphics[width= .7 \linewidth]{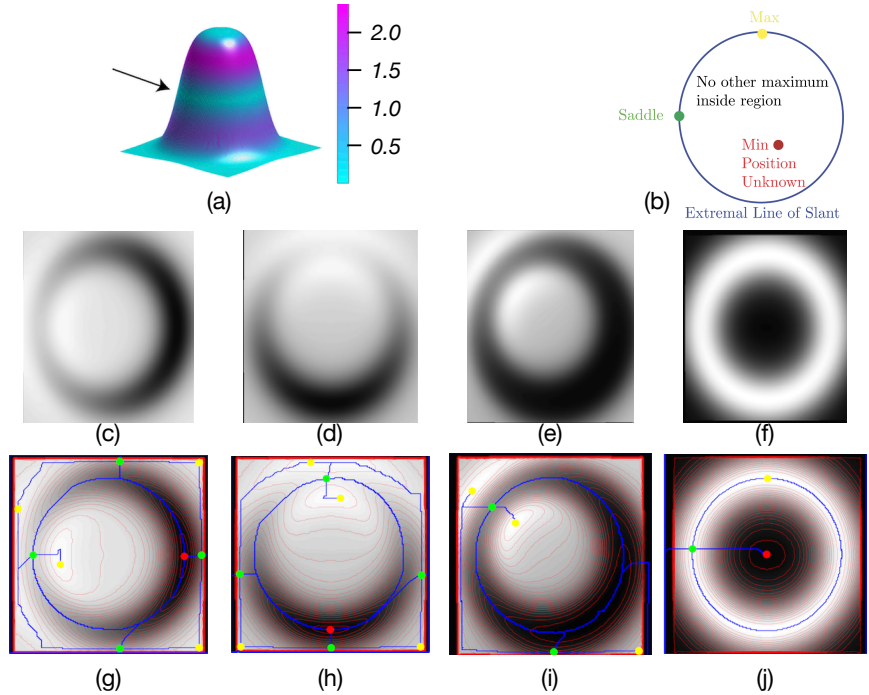}
\caption{ A sigmoidal bump on a slightly bent surface. (a) The bump is colored by the Gaussian curvature; the arrow points to the parabolic curve along which the Gaussian curvature is zero. (b) The defining template for a bump, shown in the slant domain so that an extremal curve (a critical contour in the slant domain) surrounds a minimum. (c, d) A Lambertian reflectance function on the bump illuminated from two different positions. (e) A specular rendering function. (f) The slant function for the bump viewed from above. (g - i) The isophotes and MS-complex for the three images and (j) for the slant function. 
\label{fig:sigmoid}}
\end{figure}

Several examples illustrate critical contours and their invariance properties. First consider an image of a sigmoidal bump (Fig.~\ref{fig:sigmoid}). The slant function is minimal at the top, then increases to its maximum before decreasing again. The MS-complex on the images shows how the critical contour remains invariant with respect to the light source, and agrees with the the critical contour in the slant domain that encircles the bump. There is a slant minimum (and no other maximum) inside it. The three different renderings of the sigmoidal bump each has a different intensity distribution and, of course, different isophotes. Notice how the (image) extremal contour cuts through these along a distinguished path -- it is this path that remains nearly invariant over lighting and rendering changes. 

Bumps are a special case in which the saddles merge. This case is examined in detail in \cite{kunsberg2020boundaries}, where closed critical contours in the slant domain are called extremal curves, since they are a relaxation, in a technical sense, of the occluding contour. The template for a bump illustrates a kind of definition for bumps that takes in the global nature of shape features; such global properties are not really in the domain of differential geometry.

\begin{figure}[h!]
\centering
\includegraphics[width=.7\linewidth]{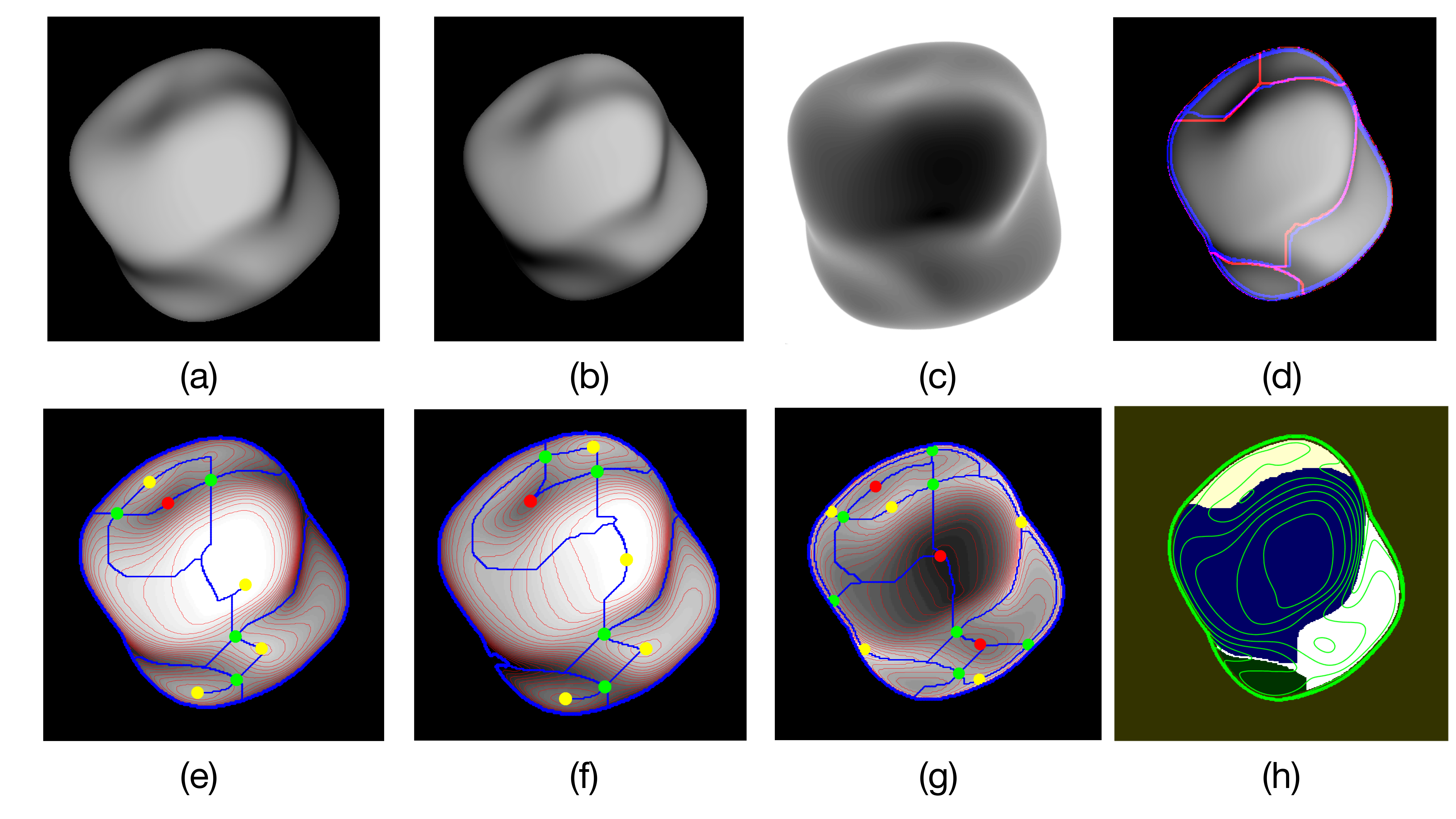}
\caption{Illustration of critical contours on a blob. (a, b) The blob shape illuminated with a Lambertian reflectance function from two different points. (c) The blob rendered by its slant. (d) The MS-complex for slant (red) and intensity (blue). Notice how they overlap almost exactly. (e, f, g) Isophotes, some critical points and MS-complex for (a, b, c) respectively. (h) Shape segmenation by critical contours.\label{fig:blob}}
\end{figure}

The next example is a blob (Fig.~\ref{fig:blob}). It is shown in Lambertian renderings from two nearby light sources and with slant as a rendering function. Even though the slant appears sort of like a photographic negative of the other images, its MS-complex and, in particular, the critical contour portions of it, remain in alignment. The sharp ridges in (a) and (b) are guaranteed to be ridges in 3D shape.

Finally, we return to the physiological example from Fig.~\ref{fig:yamane}. We show the MS-complex is invariant across those shape that most excited the neuron in IT cortex. We believe this shows that the visual system achieves a rendering-invariant representation, based on critical contours, tuned to protrusions at different relative locations. 

\begin{figure}[h!]
\begin{center}
\includegraphics[width=0.5 \linewidth]{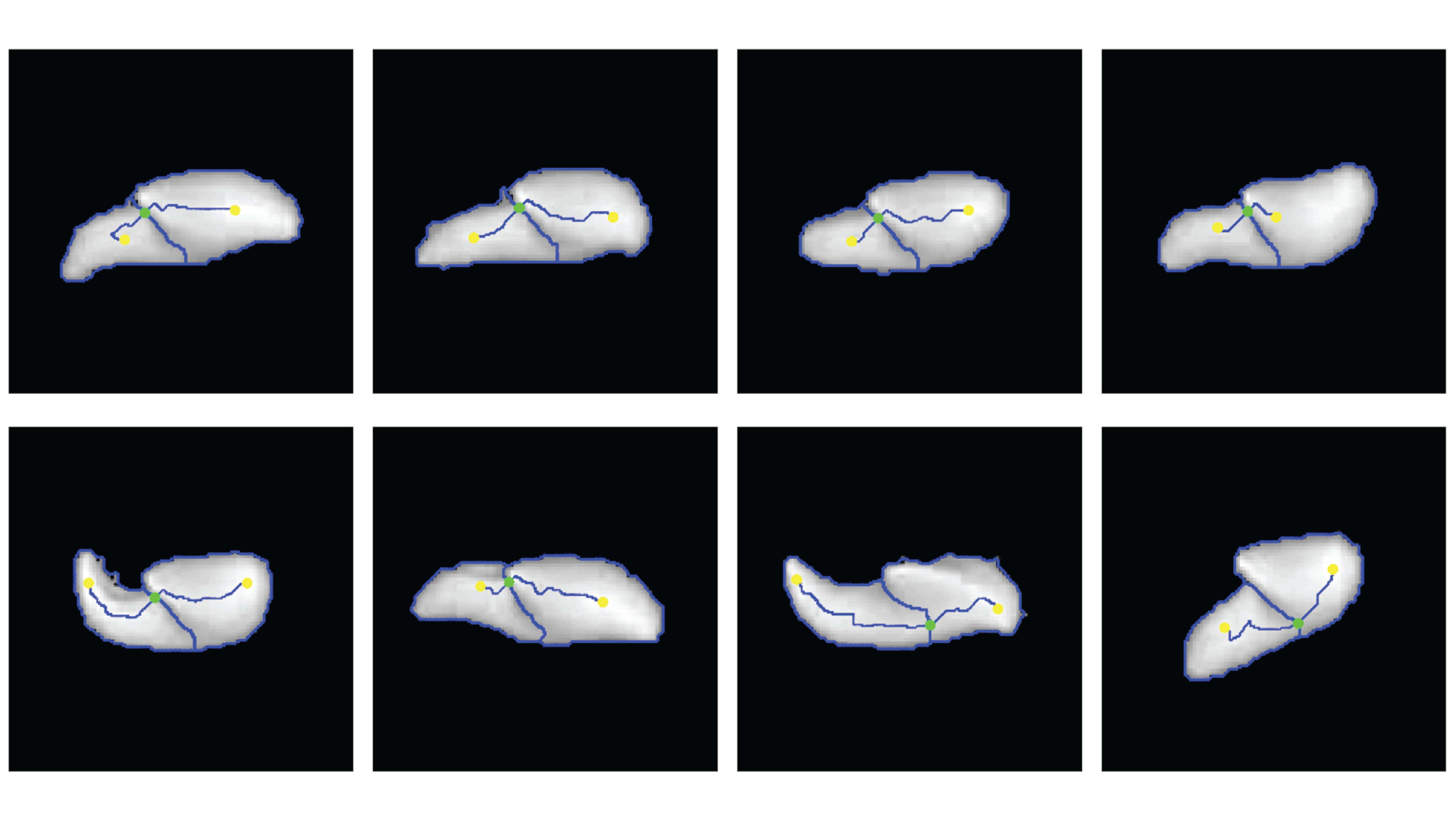} \\
\caption{\label{fig:yamane_results} Physiological evidence for critical contours.  Even though there is substantial variation in shape and pose for the most responsive stimuli in Fig.~\ref{fig:yamane},  the critical contours are topologically equivalent across them.}
\end{center}
\end{figure}

Finally, we close this section with a comment about computing critical contours. The examples shown up till now were all done using techniques from computational topology \cite{edelsbrunner2010computational, carlsson2009topology}. In this approach, discrete structures are at the base, and it enjoys persistence simplification, a globally consistent way to reveal overall structure while removing small holes and noisy details due to quantization and discretization. Specific algorithms to compute the Morse-Smale complex from discretized images have been developed by \cite{Reininghaus11, Sahner08, Weinkauf09, Weinkauf10}, among others. We used the algorithm of \cite{Reininghaus11} in our experiments.

\begin{figure}[ht!]
\begin{center}
a) \includegraphics[width= 0.6 \linewidth]{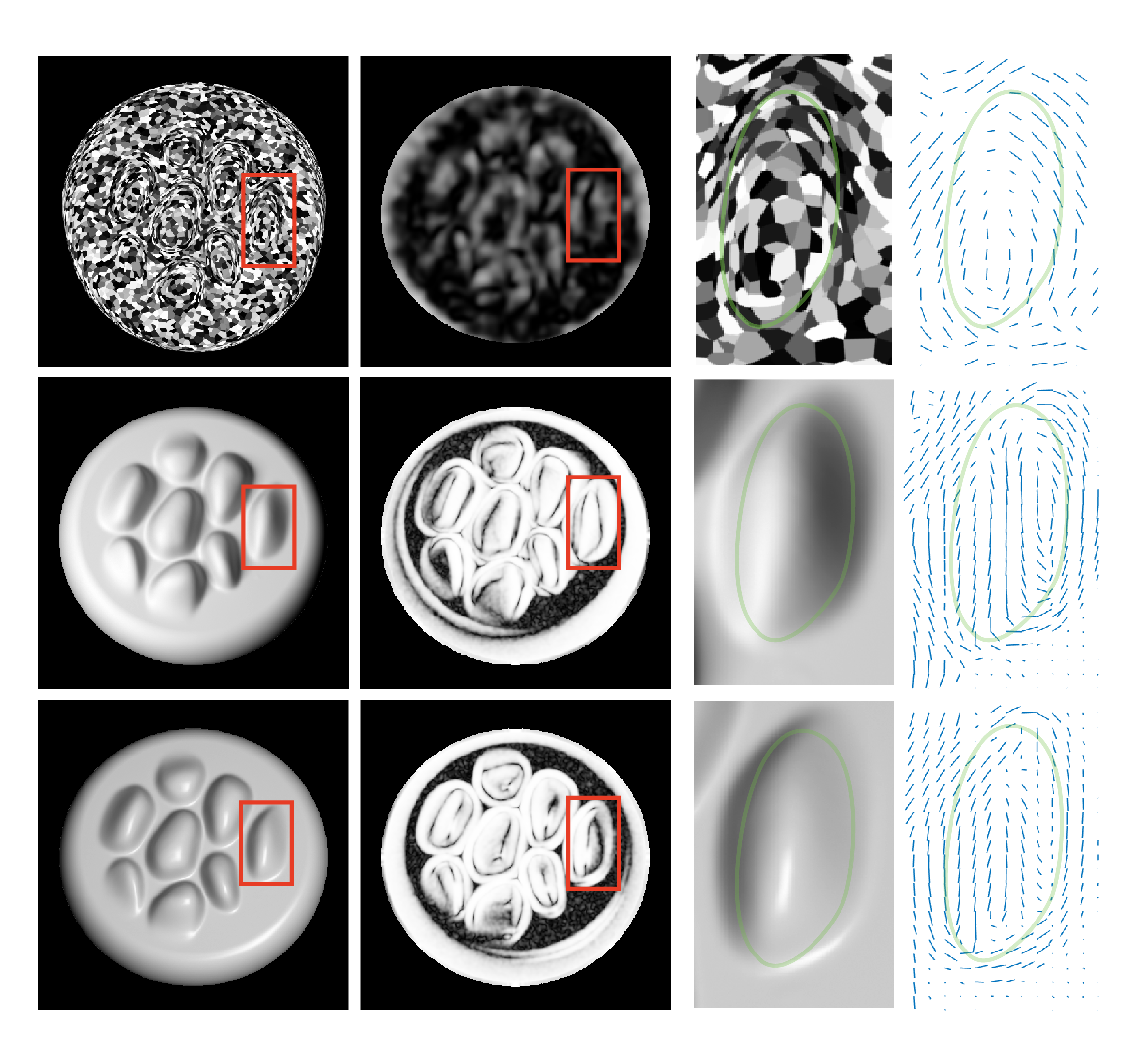}
\caption{Critical contours can be found as integral curves. Here we show  compression in shading and textured images. Across each of the rows, we show: original image, 'flow compression', an enlarged region and the 'flow direction'. Examples include a texture and two renderings of the 'pebbles' image. 
 \label{fig:texture}}
\end{center}
\end{figure}

Importantly, the structural features responsible for the MS-complex also organized the isophotes in regions of parallel flows, especially around bumps and along ridges. Perceptually we are known to be extremely sensitive to such patterns \cite{kovacs1993closed, elder1993effect} and especially oriented textures (e.g., \cite{wilson1998detection, dumoulin2007cortical, dakin2002summation}. Importantly, we are not particularly sensitive to positioning and other perturbations in them (e.g, \cite{achtman2003sensitivity}). Fig.~\ref{fig:texture} shows an example; related examples of specular reflectance functions are in \cite{kunsberg2020boundaries}.

\section{Summary and Conclusions}

We began with a brief introduction to shape-from-shading, highlighting it as an ill-posed problem and emphasizing our remarkable ability to achieve this inference. Working from a biological perspective, we began with the isophote, or shading flow structure, since this is related to the representation of image information in the early visual cortex. We then proceeded to review two threads of analysis, both inspired by the biology and tempered by a personal perspective.

In Thread I we approached shape-from-shading inferences from a differential-geometric perspective. Basically the shading flow field supported a 'moving frame' style of analysis, revealing how image derivatives were related to surface curvatures. While there was no escape from the ill-posedness, the analysis did reveal a type of characteristic organization to the solution in the neighborhood of ridges and bumps. 

Closer examination of the ridge structure led us to a second thread of analysis. This was built upon the observation in visual psychophysics that, although our perception of shape seems crisp and robust, there are actually differences between people. Despite these quantitative differences, there is substantial qualitative agreement, which we formalized in the form of critical contours. Critical contours are part of the Morse-Smale complex, a topological representation based on the critical points of smooth functions on generic surfaces. The central theorem establishes agreement between critical contours in the image and critical contours on the slant of the 3D surface. Importantly, this agreement holds across variations in lighting, material, and rendering, and is global, thereby extending the pointwise (local) desriptors previously used.

Critical contours establish a relationship between shading and contour perception, which we formalized as a concentration-of-shading limit. This leads to a novel type of prediction: if critical contours are the basis on which shape inferences are built, what about the space between them? A number of experiments are beginning to show that this space is not of primary importance; rather it is the arrangement that matters \cite{Kunsberg2018Focus, kunsberg2020boundaries}.

Many issues remain, foremost among them is that we have mainly studied the first "part" of the problem. Critical contours are only a kind of scaffold on which the shape needs to be built. Somehow it must be extended to cover the complete domain. Experiments with simple Laplacian interpolation are encouraging \cite{Kunsberg18}, but there are much more sophisticated possibilities, such as inpainting and anisotropic interpolation. While these must be developed to also be biologically plausible, this should in principle be doable. Nevertheless, the fundamential identity between critical contours in the image and critical contours on the surface remains encouraging. Perhaps this identity is why we are able to make shape inferences so robustly in the face of ill-posedness and mathematical ambiguity.

\bibliography{total-refs-3.bib}
\bibliographystyle{plain}
\end{document}


\title{Supplementary Material}
\author{Benjamin Kunsberg and Steven W. Zucker\\
\emph{Applied Mathematics, Brown University}}
\maketitle

\section{Suggestive Contours and Apparent Ridges}
Suggestive contours and apparent ridges are similar ideas, yet share two fundamental differences with our approach.  First, they are computed from the surface and there is currently no understood relationship between those surface contours and image patterns.  Second, they are defined with a pointwise condition and so do not necessarily delineate intuitive regions of the surface.  For example, there is no theory relating the suggestive contours to global surface shape and so there is no way to interpolate back the surface from suggestive contours.  Rather, we use extremal image curves as part of a larger global complex, and we show that they can be related to global surface shape.

\section{Example of Reconstruction from MS}


\begin{figure}[ht]
\begin{center}
\includegraphics[trim={1cm 2cm 7cm 2.3cm},clip, width=0.8 \linewidth]{images/complete_bunny.pdf}
\caption{These images illustrate how function values on the curves of the Morse Smale (MS) complex can be interpolated to approximate a function.  Upper left: A shaded image of the Stanford Bunny.  Center column: The MS complex at two different levels of persistence.  Right column:  Harmonic interpolation using the shading values on the MS complex and the boundary.  We see that qualitatively the images are similar to the original.  We argue similarly that the MS complex of the slant function is a good qualitative representation of the surface slant function. \label{fig:bunny}}
\end{center}
\end{figure}

We expand upon the example in Figure ?? of the Stanford Bunny.  We wish to illustrate the fact that the extremal lines of an image can, in fact, represent the majority of the percievable structure in the image. In Figure \ref{fig:inpainting}, we illustrate the process of `inpainting' an image from its extremal lines.

We start with an image of a `blob' in Fig \ref{fig:inpainting}(a). We compute the image extremal curves (the MS complex), shown in Fig \ref{fig:inpainting}(d).  These extremal curves run through the critical points of the image.  Red points correspond to minima, yellow points correspond to maxima, and green points correspond to saddles.  We now take the pixel values on the MS complex and set the vast majority of the image to constant values shown in Fig \ref{fig:inpainting}(b).  We then apply blurring via the heat kernel to get Fig \ref{fig:inpainting}(c).  We see that the quantiative pixel values on the extremal lines can `sparsely' represent the original image with nearly imperceptibly differences.  We compare the histogram of gradients of the original image and the reconstruction in Fig \ref{fig:inpainting}(e).  Here, we see that (e) is a smoother image (lower gradient norms) than (a) and so the images are different even if the qualitative perceptions are similar.

Suppose that the visual system extracted the extermal curves in (b) from the original image (a) as we are advocating.  This would not only be a sparser representation; it would also have minimal dependence on the illuminance and rendering function as shown in \cite{Kunsberg18}.

\begin{figure}[h]
\begin{center}
\includegraphics[trim={2cm 0 0cm 0},clip,width=0.8 \linewidth]{images/blur_MS_to_recreate_img.png}
\caption{a) An image of a `blob' with a vivid shape percept.  b) A `flattened' image computed from a) where the intensities on the extremal image lines are kept equivalent and, for each region, we use the mean intensity of the region.  c) A blurring of b).  d) The extremal lines and critical points of image a). e) A histogram comparing gradient norms for images c) in green, b) in blue, and a) in red. This shows that c) and a) are not  the same images although they may be perceived similarly; c) has smaller gradients than a) in general. \label{fig:inpainting}}
\end{center}
\end{figure}

\section{Generic Argument against the Twisted Surface}

We remarked in Section 3.1 that there was a generic argument that the normal field along a slant extremal curve must uniformly `point to one side'.  We now give this argument, following Freeman's approach in \cite{freeman94}.  The gist of the argument is one can define a probability distribution on solutions to a constraint based on the stability of `nuisance parameters'.  Here, we take the nuisance parameter to be the direction of illumination.  The most generic solutions then will be the solutions that are most stable with respect to changes in the illumination.

Define, using the notation in \cite{freeman94},
\begin{align}
\bm{\beta} & = \D \n \\
\bm{Y} & =  \D I 
\end{align}

Now, consider the following rendering function $\mathbf{g}$:

\begin{align}
\mathbf{g}(\l, \bm{\beta} ) & = \bm{\hat{Y}} \\
& = \l \tr \bm{\beta}
\end{align}

Following \cite{freeman94, Freeman96}, we call $\l$ the \emph{generic variable}, $\bm{\beta}$ the \emph{scene parameters}, and $\bm{Y}$ the \emph{observations}.  $\l$ is an unknown vector in $\mathbb{R}^3$.  Following \cite{freeman94, Freeman96}, we assume a Gaussian noise model on the observations $\bm{Y}$: 
\begin{align}
\bm{Y} = \hat{\bm{Y}} + \bm{T}
\end{align}
 
\noindent
where $\hat{\bm{Y}}$ is the ideal rendered observation and $T \sim N(0, \bm{\Sigma})$ with $\bm{\Sigma} = \text{diag}(\sigma^2)$ for some $\sigma \in \mathbb{R}^+$.  For the noise model, we have

\begin{align}
P(\bm{Y} | \bm{\beta}, \l) & =  \frac{1}{(\sqrt{2 \pi \sigma^2})^m} e^{- \frac{ || \bm{Y} -\mathbf{g}(\l, \bm{\beta}) ||^2}{2 \sigma^2}}
\end{align}

\noindent
This is the probability of the surface $\D \n$ given a noisy observation of the image gradient $\D I$.  Applying Bayes' theorem and integrating over the nuisance variable $\l$ yields the posterior distribution:

\begin{align}
P(\bm{\beta} | \bm{Y}) & = k \exp \left( \frac{ - ||\bm{Y} - \mathbf{g}(\l_0 , \bm{\beta}) ||^2}{2 \sigma^2} \right) \nonumber \\
& \cdot [ P_{\bm{\beta}} (\bm{\beta}) P_{\l} (\l_0)] \frac{1}{\sqrt{\det (\bm{A})}} \label{eqn:posterior}\\
& = k \hspace{2mm} \text{(fidelity)} \nonumber \\
& \cdot \hspace{2mm} \text{(prior probability)} \hspace{2mm} \text{(genericity)} \nonumber
\end{align}
\noindent
where $P_{\bm{\beta}} (\bm{\beta}), P_{\l} (\l_0)$ are prior distributions on the surface and light source parameters, $\l_0$ is the light source that can best account for the observations given a chosen $\bm{\beta}$ and $\bm{A}$ is a matrix with the following elements:

\begin{equation}
A_{ij} = \mathbf{g}'_i \cdot \mathbf{g}'_j - (\bm{Y} - \mathbf{g}(\l_0, \bm{\beta})) \cdot \mathbf{g}''_{ij}
\end{equation}
with 
\begin{align}
\mathbf{g}'_i & = \frac{\partial \mathbf{g} (\l, \bm{\beta})}{\partial l_i} \Big|_{\l = \l_0} \\
\mathbf{g}''_{ij} & = \frac{\partial^2 \mathbf{g} (\l, \bm{\beta})}{\partial l_i \partial l_j} \Big|_{\l = \l_0}
\end{align}

For more details of the previous Bayesian analysis, consult \cite{Freeman96}. We now seek the solutions that maximize the posterior probability $P(\bm{\beta} | \bm{Y})$.  From (\ref{eqn:posterior}), we maximize by choosing $\bm{\beta}$ and $\l_0$ so that $||\bm{Y} - \mathbf{g}(\l_0 , \bm{\beta}) || = 0$ while at the same time setting $\det(\bm{A}) = 0$.  

This means we can maximize the probability of the surface property $\D \n$  given the image gradient $\D I$ if we set $\det(\bm{A}) = 0$.  This is equivalent to ensuring that $\D \n$ is low rank and so the surface has zero (or minimal) Gaussian curvature in the given patch.  As shown in Fig ?? in the main paper, we see that the Gaussian curvature is very large for the twisted surface whereas the Gaussian image is 0 for the non-twisted surface.


\section{Depth Functions for Figure \ref{fig:spherical_ridge}}

We show the height functions for the different labeling cases in Figure ??a and Figure ??d in the main paper.
\begin{figure}[h]
\begin{center}
\includegraphics[width= 0.4 \linewidth]{images/good_ridge2.png}
\includegraphics[width= 0.4 \linewidth]{images/alternate_ridge2.png}
\caption{Left: Surface corresponding to the topology described in labeling (a) in Fig \ref{fig:spherical_ridge}. Right: Surface corresponding to the topology described in labeling (d) in Fig \ref{fig:spherical_ridge}.  Note that diagrams b) and c) are the negative height functions of a) and d).  Left is more generic than Right.  You can think of Right as starting with Left and pulling the right side of the surface up until it is above the ridge.}
\end{center}
\end{figure}

\section{The Dual Ridge example}

We can extrapolate the ideas illustrated in Fig ?? to multiple ridges.  Here is an example of the extremal curves for a `two-ridge spherical coin'.  The possible labelings can clearly be deduced in the same manner as Fig ??.   

\begin{figure}[h]
\begin{center}
\includegraphics[width=1 \linewidth]{images/two_ridge_example.png}
\caption{Left: An image of the `two-ridge spherical coin.'  Center: The computed extremal image curves.  Right: The author's perceived labeling.}
\end{center}
\end{figure}

\newpage

\printbibliography